\documentclass[10pt,journal,compsoc]{IEEEtran}
\ifCLASSOPTIONcompsoc
  \usepackage[nocompress]{cite}
\else
  \usepackage{cite}
\fi

\usepackage{url}
\usepackage{graphicx}
\usepackage{amsmath} 
\usepackage{bm}
\usepackage{amsfonts,amssymb} 
\usepackage{booktabs}
\usepackage{bbding}
\usepackage{makecell}
\usepackage{xcolor}
\usepackage{wasysym}
\usepackage{multirow}
\usepackage[pagebackref=false,breaklinks=true,colorlinks,bookmarks=false,urlcolor=magenta]{hyperref}
\newcommand{\fr}[1]{\textcolor{black}{#1}} 
\begin{document}

\title{Robust Visual Question Answering: Datasets, Methods, and Future Challenges}

\author{Jie Ma,~\IEEEmembership{Member,~IEEE,}
        Pinghui Wang, ~\IEEEmembership{Senior Member,~IEEE,}
        Dechen Kong\textsuperscript{\dag}, 
        Zewei Wang\textsuperscript{\dag},
        Jun Liu, ~\IEEEmembership{Senior Member,~IEEE},
        Hongbin Pei,
        and
        Junzhou Zhao
\IEEEcompsocitemizethanks{
\IEEEcompsocthanksitem Jie Ma, Pinghui Wang, Hongbin Pei, Junzhou Zhao, and Jia Di are with the  Ministry of Education of Key Laboratory for Intelligent Networks and Network Security, School of Cyber Science and Engineering, Xi’an Jiaotong University, Xi’an, Shaanxi 710049, China. 
\IEEEcompsocthanksitem Dechen Kong and Zewei Wang are with the  Ministry of Education of Key Laboratory for Intelligent Networks and Network Security, School of Automation Science and Engineering, Xi'an Jiaotong University, Xi'an, Shaanxi 710049, China.
\IEEEcompsocthanksitem Jun Liu is with the Shannxi Provincial Key Laboratory of Big Data Knowledge Engineering, School of Computer Science and Technology, Xi'an Jiaotong University, Xi'an, Shaanxi 710049, China.
\IEEEcompsocthanksitem Dechen Kong and Zewei Wang contribute equally to the work. 
\IEEEcompsocthanksitem Pinghui Wang is the corresponding author.
}
}

\markboth{IEEE Transactions on Pattern Analysis and Machine Intelligence}
{Ma \MakeLowercase{\textit{et al.}}: Robust Visual Question Answering: Datasets, Methods, and Future Challenges}


\IEEEtitleabstractindextext{
\begin{abstract}
Visual question answering requires a system to provide an accurate natural language answer given an image and a natural language question. However, it is widely recognized that previous generic VQA methods often tend to memorize biases present in the training data rather than learning proper behaviors, such as grounding images before predicting answers. Therefore, these methods usually achieve high in-distribution but poor out-of-distribution performance. In recent years, various datasets and debiasing methods have been proposed to evaluate and enhance the VQA robustness, respectively. This paper provides the first comprehensive survey focused on this emerging fashion. Specifically, we first provide an overview of the development process of datasets from in-distribution and out-of-distribution perspectives. Then, we examine the evaluation metrics employed by these datasets. Thirdly, we propose a typology that presents the development process, similarities and differences, robustness comparison, and technical features of existing debiasing methods. Furthermore, we analyze and discuss the robustness of representative vision-and-language pre-training models on VQA. Finally, through a thorough review of the available literature and experimental analysis, we discuss the key areas for future research from various viewpoints. 

\end{abstract}

}
\maketitle
\IEEEdisplaynontitleabstractindextext
\IEEEpeerreviewmaketitle

\IEEEraisesectionheading{\section{Introduction}}
\IEEEPARstart{V}{isual} Question Answering (VQA) aims to build intelligent machines that can provide a natural language answer accurately given an image and a natural language question about the image~\cite{antol2015vqa}. The goal of VQA, which bridges computer vision and natural language processing, is to teach machines to see and read simultaneously like humans. This task exhibits a multitude of applications, encompassing areas such as providing blind and visually impaired individuals with information about the surrounding world, facilitating image retrieval in the absence of metadata \cite{kafle2017visual}, empowering intelligent virtual assistants \cite{liu2023older}, enabling visual recommendation systems \cite{zangerle2022evaluating}, and contributing to autonomous driving \cite{chen2022milestones}. For instance, we can use the VQA approach to query ``Is there a panda in the image?" across all candidate images to identify those that contain pandas.
\begin{figure*}[t]
	\centering  
	\includegraphics[width=0.95\textwidth]{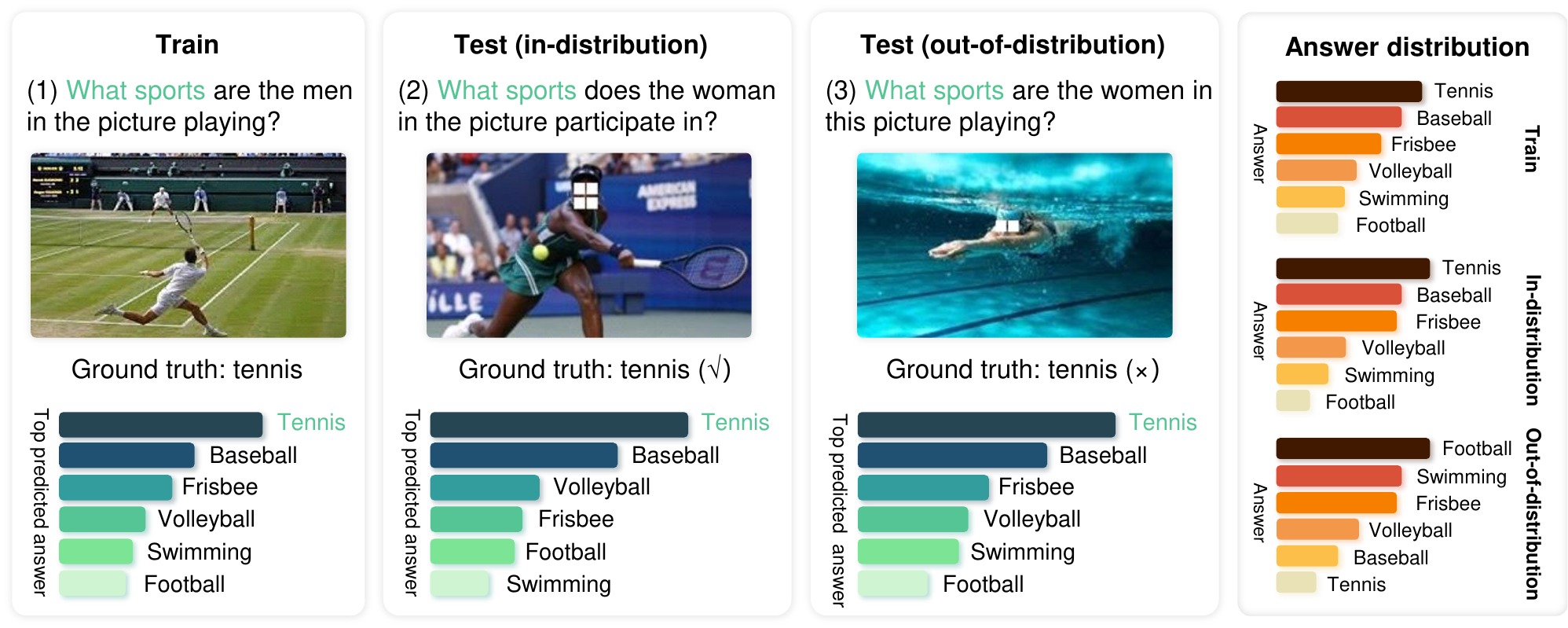}
	\caption{Illustration of generic VQA methods in the in-distribution and out-of-distribution test scenarios. They predict answers by learning strong language bias, such as the connections between critical words ``what", and ``sports" in questions and the most frequent answer ``tennis", rather than grounding images, which results in their high In-Distribution (ID) but poor Out-Of-Distribution (OOD) test performance. The ID situation refers to scenarios where the distribution is similar to that of the corresponding training split. The OOD scenario, on the other hand, applies to cases where the distribution differs or even opposes that of the training split.}
	\label{fig:vqa-example}
\end{figure*}

In the last few years, VQA has witnessed extensive exploration across various domains, encompassing scenarios involving real natural images \cite{antol2015vqa}, synthetic images \cite{johnson2017clevr}, scientific charts \cite{siegel2016figureseer}, and diagrams \cite{ma2023diagram}. Notably, VQA for real natural images, which is the primary focus of this paper, is the most widely studied among the mentioned domains. It has garnered significant attention in the research community \cite{park2021bridge,rahman2021improved,cao2022bilateral}, which can be attributed to two key developments. First, several datasets, such as VQA v1 \cite{antol2015vqa}, VQA v2 \cite{goyal2017making} for fine-grained detection and recognition, FVQA \cite{wang2017fvqa}, OK-VQA \cite{marino2019ok} for reasoning based on external knowledge, and GQA for compositional reasoning \cite{hudson2019gqa}, have been built to evaluate the ability of methods from different views. Second, a variety of VQA methods have been proposed, which can be classified into three groups \cite{kafle2017visual}: joint embedding, attention mechanism, and external knowledge. Joint embedding-based methods \cite{fukui2016multimodal, ma2016learning, antol2015vqa}, in particular, project image and question representations into a common space to predict answers. These methods typically learn coarse-grained multi-modal representations, which may bring noisy or irrelevant information into the prediction stage. To address this issue, attention mechanism-based methods \cite{yang2016stacked, anderson2018bottom, cadene2019murel} usually fuse the representations of questions and images based on the learned importance of objects and words. In the real-world scenario, VQA often requires machines to understand not only image contents but also external prior information that ranges from common sense to encyclopedic knowledge. For instance, considering the question ``How many fungal plants are there in the picture?", the machine should comprehend the word ``fungal" and know which plant belongs to this category. To this end, external knowledge-based methods \cite{wu2022multi, ding2022mukea, Ravi}, which have two popular lines: traditional information retrieving and large language model prompting, aim to utilize the relevant knowledge produced from an external source to address the knowledge-based VQA. The former links the knowledge retrieved from large-scale knowledge bases such as DBpedia \cite{auer2007dbpedia}, Freebase \cite{bollacker2008}, and YAGO \cite{hoffart2013yago2}, into multimodality learning \cite{8269806}. The latter \cite{guo2023images, shao2023prompting, gao2022transform} first employs an off-the-shelf captioning model to translate images into a caption and then integrates questions, captions, and a few in-context examples to induce the large language model such as GPT-3 to predict answers.

However, in parallel with the above works, several studies \cite{agrawal2018don,niu2021introspective,Kim2023CVPR,zhang2023next,song2023recovering} found that the aforementioned generic methods tend to memorize statistical regularities or bias in the training data rather than ground images to predict answers. For example, in the middle bar chart of the fourth column of Fig. \ref{fig:vqa-example}, we can see that ``tennis" is the most frequent answer. These methods answer the questions of the second column mainly by exploiting the connections between critical words ``what", and ``sports" of the questions and ``tennis". This will cause these methods to perform well in the In-Distribution (ID) test scenario that has similar answer distributions with the training split, such as the distribution in the middle bar chart of the fourth column, but poorly in the Out-Of-Distribution (OOD) test situation that has different or even reversed answer distributions, such as the distribution in the bottom bar chart. To address this issue, a significant body of literature on VQA has emerged in recent years, with a particular focus on eliminating bias \cite{agrawal2018don, cadene2019rubi, gupta2022swapmix} and evaluating robustness \cite{agrawal2018don, kervadec2021roses, si2022language}. 

The purpose of this paper is to provide a comprehensive and systematic overview of the methods, datasets, and future challenges in the area of robust VQA. To the best of our knowledge, this paper is the first survey on the mentioned topic, although there exist some surveys \cite{kafle2017visual,de2023visual,kafle2019challenges, bernardi2021linguistic} on VQA. In Section \ref{sec:backg}, we establish preliminary concepts for generic and robust VQA. Section \ref{sec:data} discusses the datasets from various perspectives including constructions, image sources, amounts of images and questions, and focuses. They are classified into two categories based on the ID and OOD settings. Section \ref{sec:eval} reviews the evaluation metrics employed in the mentioned datasets including single and composite metrics. In Section \ref{sec:meth}, we undertake a critical analysis of debiasing methods, categorizing them into four classes: ensemble learning, data augmentation, self-supervised contrastive learning, and answer re-ranking. Section \ref{sec:VLM} reviews the development of representative vision-and-language pre-training methods and divides them into four classes based on the relative computational size of text encoders, image encoders, and modality interaction modules. We also discuss the robustness of these methods on the most commonly used VQA v2 \cite{goyal2017making} (ID) and VQA-CP (OOD) \cite{agrawal2018don} datasets. Furthermore, in Section \ref{sec:dis}, we conduct in-depth discussions of future challenges from the perspective of improvements of annotation qualities, ongoing developments in dataset creation, evaluation metric advancements, method robustness, and robustness assessments, drawing on our experimental results and literature overview. Finally, we present our concluding remarks in Section \ref{sec:con}.

\section{Preliminaries}
\label{sec:backg}
In this section, we first describe the task formulation of VQA. Then, we briefly introduce the paradigm of current VQA methods. Finally, we define robust VQA methods from the perspective of debiasing.

\subsection{Task Formulation}
Since the VQA task was proposed, it was initially treated as a discriminative task. In recent years, as large models have rapidly advanced, there has been a growing body of research that considers VQA as a generative task.

\noindent\textbf{Discriminative VQA} is formulated as a classification task. Given a dataset $\mathcal{D}$ consisting of $n$ triplets $\{(v_i, q_i, a_i)\}_{i=1}^{n}$ with an image $v_i \in \mathcal{V}$, a question $q_i \in \mathcal{Q}$, and an answer $a_i \in \mathcal{A}$, discriminative VQA requires machines to optimize the parameters $\theta^{(\mathrm{d})}$ and predict answers $\hat{a}_i^{(\mathrm{d})}$ accurately:
\begin{equation}
    \hat{a}_i^{(\mathrm{d})} = \mathop{\arg\max}_{a_i \in \mathcal{A}} p(a_i | v_i, q_i; \theta^{(\mathrm{d})}).
\end{equation}

\noindent\textbf{Generative VQA} is formulated as a generation task. Given a dataset $\mathcal{D}$, generative VQA requires machines to predict answers token by token, where $\hat{y}_j$ is the $j_{th}$ token of the predicted answer $\hat{a}_i^{(\mathrm{g})}$. $\hat{y}_j$ is obtained by optimizing the parameters $\theta^{(\mathrm{g})}$ to maximize the conditional probability $p(y_j | (\hat{y}_1, ..., \hat{y}_{j-1}), v_i, q_i; \theta^{(\mathrm{g})})$:
\begin{equation}
    \begin{split}
        & \hat{a}_i^{(\mathrm{g})} =  (\hat{y}_1, ..., \hat{y}_k), \\
        & \hat{y}_1 = \mathop{\arg\max}_{y_1 \in \mathcal{Y}} p(y_1 | (v_i, q_i; \theta^{(\mathrm{g})}), \\
        & \hat{y}_j = \mathop{\arg\max}_{y_j \in \mathcal{Y}} \prod_{j=2}^{k} p(y_j | (\hat{y}_1, ..., \hat{y}_{j-1}), v_i, q_i; \theta^{(\mathrm{g})}),
    \end{split}
\end{equation}
where $\mathcal{Y}$ represents the set of all tokens in the corpus and $k$ represents the number of tokens in $\hat{a}_i$.

\subsection{VQA Paradigm}
Existing debiasing (robust) methods typically utilize non-debiasing methods as a foundational backbone or module. As a result, we will present the paradigm from two distinct perspectives: non-debiasing and debiasing.
\subsubsection{Non-debiasing}
\noindent\textbf{Discriminative VQA} methods \cite{fukui2016multimodal,kafle2017analysis,wu2017image,anderson2018bottom,cadene2019murel,guo2021re} typically leverage an image encoder $E_{\mathrm{v}}: \mathcal{V} \rightarrow \mathbb{R}^{n_{\mathrm{v}} \times d_{\mathrm{v}}}$ to learn $n_{\mathrm{v}}$ region-level (patch-level) representations with $d_{\mathrm{v}}$ dimensions, a question encoder $E_{\mathrm{q}}: \mathcal{Q} \rightarrow \mathbb{R}^{n_{\mathrm{q}} \times d_{\mathrm{q}}}$ to output $n_{\mathrm{q}}$ word-level vectors with $d_{\mathrm{q}}$ dimensions, a multi-modality encoder $E_{\mathrm{m}}: \mathbb{R}^{n_{\mathrm{q}} \times d_{\mathrm{q}}} \times \mathbb{R}^{n_{\mathrm{v}} \times d_{\mathrm{v}}} \rightarrow \mathbb{R}^{d_{\mathrm{m}}}$ to learn fusion representations with $d_{\mathrm{m}}$ dimensions, and a classifier $E_{\mathrm{c}}: \mathbb{R}^{d_{\mathrm{m}}} \rightarrow \mathbb{R}^{|\mathcal{A}|}$ to obtain predictions over the answer space $\mathcal{A}$. The paradigm can be denoted as follows:
\begin{figure*}[t]
	\centering  
	\includegraphics[width=0.95\textwidth]{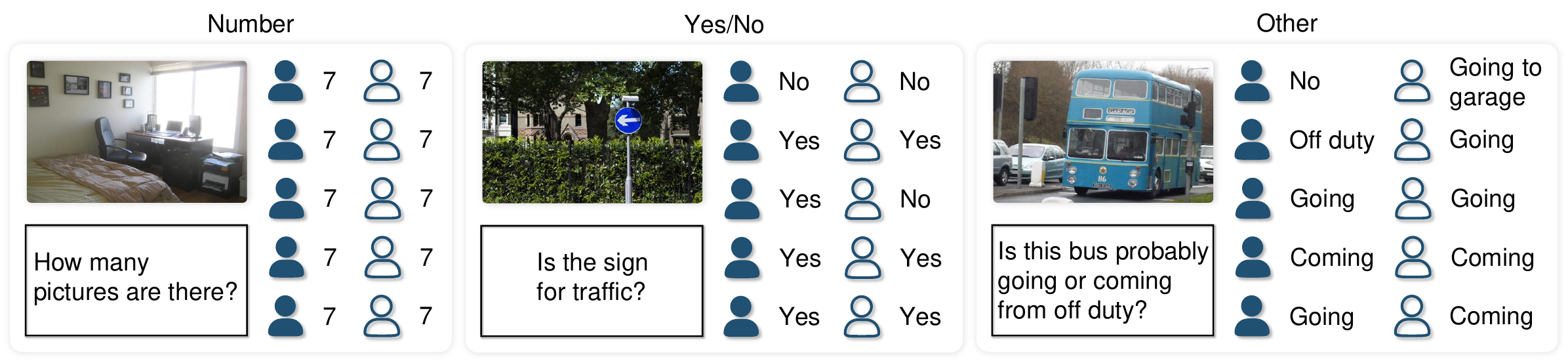}
	\caption{VQA v1 example. Each question in this dataset is annotated by ten humans. Therefore, a question may have multiple different ground-truth answers. The dataset is classified into three categories according to answer types: ``Number", ``Yes/No", and ``Other".}
	\label{fig:vqa-v1-eg}
\end{figure*}
\begin{equation}
    \hat{a}_i^{\mathrm{(d)}} = f^{\mathrm{(d)}}(v_i, q_i; \theta^{\mathrm{(d)}}) = E_{\mathrm{c}} ( E_{\mathrm{m}} ( E_{\mathrm{v}}(v_i), E_{\mathrm{q}}(q_i) ) ) . \label{eq:gp}
\end{equation}

Recently, the success \cite{wolf2020transformers,naseem2021semantics,garg2020tanda,glass2020span,zhang2020sg,zhou2021topicbert} of Transformer in natural language processing has given rise to a new pattern for VQA. A variety of studies \cite{li2019visualbert,chen2020uniter,liDFGJ20} leverage the Transformer to replace $E_{\mathrm{q}}$, $E_{\mathrm{v}}$ and $E_{\mathrm{m}}$. Due to the unique nature of the Transformer, there are two approaches to accomplishing this objective: dual stream and single stream. The former employs different Transformers to replace the encoder in Eq. \eqref{eq:gp}, respectively. In contrast, the latter regards the input modalities as tokens with the same type and utilizes transformers $E_{\mathrm{t}}$ to learn their joint representations:
\begin{equation}
    \hat{a}_i^{(\mathrm{d})} = f^{\mathrm{(d)}}(v_i, q_i; \theta^{(\mathrm{d})}) = E_{\mathrm{c}}(E_{\mathrm{t}}(\bm{v}_i||\bm{q}_i)), \label{eq:os}
\end{equation}
where $\bm{v}_i$ is the representation of regions or patches, $\bm{q}_i$ denotes the word embeddings, and $||$ represents the concatenation. It is worth noting that single- and dual-stream encodings also exist in the encoding stage of the generative VQA paradigm.


\noindent\textbf{Generative VQA} methods typically employ a question encoder $E_{\mathrm{q}}: \mathcal{Q} \rightarrow \mathbb{R}^{n_{\mathrm{q}} \times d_{\mathrm{q}}}$ to transform the question into $n_{\mathrm{q}}$ word-level embeddings with $d_{\mathrm{q}}$ dimensions and an image encoder $E_{\mathrm{v}}: \mathcal{V} \rightarrow \mathbb{R}^{n_{\mathrm{v}} \times d_{\mathrm{v}}}$ to learn $n_{\mathrm{v}}$ region-level (patch-level) representations with $d_{\mathrm{v}}$ dimensions. These representations serve as the foundation for generating answers. Then, they leverage a decoder module $\mathrm{Decoder}: \mathbb{R}^{d_{\mathrm{q}}} \times \mathbb{R}^{d_{\mathrm{v}}} \rightarrow \mathbb{R}^{|\mathcal{\mathcal{Y}}|}$ to generate answers from the fused multimodality representations. Mathematically, this generative paradigm can be expressed as follows:
\begin{equation}
\hat{a}_i^{\mathrm{(g)}} = f^{\mathrm{(g)}}(q_i, v_i; \theta^{\mathrm{(g)}}) = \mathrm{Decoder}(E_{\mathrm{q}}(q_i), E_{\mathrm{v}}(v_i)). \label{eq:gen-vqa}
\end{equation}

Generative VQA methods are known for their capacity to produce diverse and contextually relevant answers, which renders them highly attractive for tasks that necessitate creative responses and natural language generation. Their adaptability to open-ended questions and the potential for richer interactions with users make them a distinct and valuable component of the VQA landscape. However, it is essential to acknowledge that, despite their strengths, the generative VQA paradigm has not been extensively delved into the realm of robust VQA.

\subsubsection{Debiasing}
Recent studies \cite{agrawal2018don,cadene2019rubi,niu2021introspective,gupta2022swapmix} found that the non-debiasing methods are apt to exploit the training bias excessively rather than learn proper behaviors. For example, they may learn the spurious connections between critical words of questions and answers, such as ``what", ``sports" and ``tennis". In other words, the non-debiasing methods predict answers mainly relying on one of $E_{\mathrm{v}}(v_i)$ and $E_{\mathrm{q}}(q_i)$. This can be formulated as follows:
\begin{align}
    & \hat{a}_{i}^{(\mathrm{l})} = f^{\mathrm{(l)}}(v_i, q_i; \theta_{\mathrm{l}}) = E_{\mathrm{c}}(E_{\mathrm{q}}(q_i)), \\
    & \hat{a}_{i}^{(\mathrm{v})} = f^{\mathrm{(v)}}(v_i, q_i; \theta_{\mathrm{v}}) = E_{\mathrm{c}}(E_{\mathrm{v}}(v_i)),
\end{align}
where $f^{\mathrm{(l)}}$ refers to the language bias learning, $f^{\mathrm{(v)}}$ represents the vision bias learning, $\hat{a}_{i}^{\mathrm{(l)}}$ refers to the predicted answer relying on the language bias learning, and $\hat{a}_{i}^{\mathrm{(v)}}$ refers to the predicted answer relying on the vision bias learning. According to the investigation in \cite{agrawal2018don, si2022language}, it has been found that language bias learning is more prevalent than vision bias learning. Recent studies \cite{dancette2021beyond,si2022language} have found that there also exists some possibility to predict answers $\hat{a}_{i}^{(\mathrm{m})}$ using multimodal bias learning. This can be expressed as follows.
\begin{align}
    \hat{a}_{i}^{(\mathrm{m})} = f^{\mathrm{(m)}}(v_i, q_i; \theta^{(\mathrm{m})}).
\end{align}

The mentioned bias learning will cause the non-debiasing methods to obtain high ID but poor OOD performance. In other words, the methods are incapable of dealing with various changing real-world situations. In comparison, based on non-debiasing methods, debiasing methods aim at mitigating bias learning to achieve robust VQA performance in both scenarios. Existing debiasing methods primarily focus on eliminating the impact of $f^{\mathrm{(l)}}$, $f^{\mathrm{(v)}}$ or $f^{\mathrm{(m)}}$ on the $f$ described in Eq. \eqref{eq:gp}, \eqref{eq:os} and \eqref{eq:gen-vqa}. The impact is always mitigated in explicit or implicit ways. Specifically, the answer prediction like $\hat{a}_i^{\mathrm{(d)}}$ is explicitly changed in the former methods. For example, ensemble learning-based methods discussed in Section \ref{sec:meth} combine $\hat{a}_i^{\mathrm{(d)}}$ and biased prediction such as $\hat{a}_{i}^{(\mathrm{l})}$ to overcome biases. In contrast, $\hat{a}_i^{\mathrm{(d)}}$ can also be affected in an implicit manner. For instance, data augmentation-based methods construct additional samples to drive $f^{\mathrm{(d)}}$ and $f^{\mathrm{(g)}}$ to eliminate biases implicitly. We will delve into a detailed discussion of how debiasing methods address biases in Section \ref{sec:meth}.

\section{Datasets}
\label{sec:data}
A substantial volume of VQA datasets has been constructed encompassing diverse perspectives. This section presents a comprehensive review of these datasets, categorizing them into two groups according to their ID and OOD settings.
\subsection{In-Distribution Setting}
\emph{ID settings typically represent that the test split's distribution is similar to that of the corresponding training split.} A variety of ID datasets \cite{antol2015vqa,goyal2017making,gurari2018vizwiz,hudson2019gqa,nguyen2023vlsp,changpinyo2022towards,pfeiffer2022xgqa} have been developed from different angles and have made significant contributions to robust VQA research. These datasets can be particularly useful in evaluating the effectiveness of VQA methods, as they are specifically designed to assess various capabilities such as fine-grained detection and recognition, multilingual ability, as well as common sense reasoning.

\noindent\textbf{VQA v1}~\cite{antol2015vqa}. This dataset was developed by asking ten human subjects the same question about a real image. As shown in the ``other" question of Fig. \ref{fig:vqa-v1-eg}, we can see that ten annotators produce five gold answers, including ``No", ``Off duty", ``Going", ``Coming" and ``Going to garage". Therefore, the question in this dataset may have several ground-truth answers. VQA v1 is the first large-scale, open-ended, and free-form dataset to assess the ability of VQA methods such as fine-grained detection \cite{yang2018learning,zhao2023learning,zha2019context}, recognition \cite{zhao2020exploring,xie2020adversarial,chen2023federated}, and counting \cite{acharya2019tallyqa,sindagi2019ha}. However, Goyal \textit{et al.} \cite{goyal2017making} found that VQA v1 contains strong language priors or bias. For example, the most common answer ``tennis" is the correct answer for 41\% of the questions starting with ``what sport is". This may drive VQA methods to answer questions depending on the bias rather than grounding images.

\begin{figure}[t]
	\centering  
	\includegraphics[width=\columnwidth]{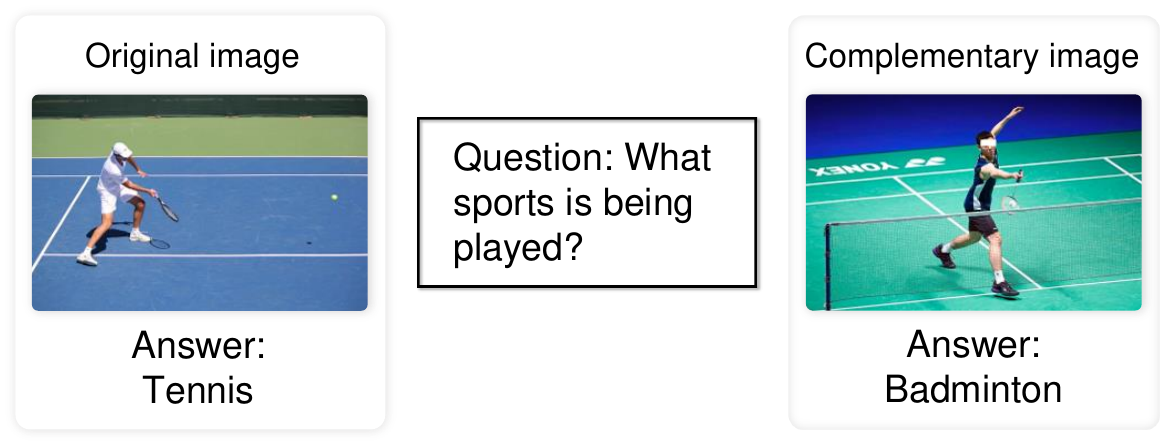}
	\caption{Illustration of balancing answer distributions in VQA v2. This dataset incorporates a complementary image that pertains to the same question in VQA v1 but has a different answer.}
	\label{fig:vqa-v2-eg}
\end{figure}

\noindent\textbf{VQA v2}~\cite{goyal2017making}. To alleviate the mentioned dilemma, this dataset was developed by collecting complementary images for almost every question in VQA v1. Specifically, the two images appear to be similar but have different answers to the question. As shown in Fig. \ref{fig:vqa-v2-eg}, VQA v2 adds a complementary image for the question ``What sport is being played?" to mitigate the bias in VQA v1, where ``badminton" is the complemented answer.

Despite achieving substantial success \cite{jiang2020defense,li2019relation,9525040,nguyen2022coarse,luo2022towards}, VQA v2 still faces two issues \cite{rosenberg2021vqa, zhang2016yin, das2017human, johnson2017clevr}. Firstly, the similarity between the distributions of training and test splits still exists. The construction of splits in the aforementioned dataset is based on the corresponding partitions of MS COCO. It is worth noting that the partitions of MS COCO may already exhibit similar distributions, which unavoidably influences the distribution of VQA v1 and v2. Secondly, the lack of meaningful categorization for questions in existing datasets poses a challenge in comparing the performance of individual algorithms.

\noindent\textbf{TDIUC}~\cite{kafle2017analysis}. To address the mentioned issues, this dataset was developed by importing questions from COCO-based VQA \cite{malinowski2014multi,antol2015vqa,goyal2017making} and Visual Genome \cite{krishna2017visual}. These questions are divided into 12 types such as ``positional reasoning" and ``activity recognition". However, the simplicity of the question in the aforementioned dataset exacerbates the limitation stemming from the models' dependence on statistical biases. Furthermore, the lack of annotations regarding question structures and contents increases the likelihood that models will rely on statistical regularities during training.

\noindent\textbf{GQA}~\cite{hudson2019gqa}. To address the mentioned issues, this dataset was developed by leveraging the content provided by the scene graphs in Visual Genome \cite{krishna2017visual}, such as object information, attributes and relations, and structures from extensive linguistic grammar. Therefore, the question is more complex compared with that in VQA v2. This can decrease the reliance on training bias in generating answers.

\noindent\textbf{COVR}~\cite{bogin2021covr}. Similar to GQA, this dataset was developed by extracting subgraphs of similar but distracting images within Visual Genome. Unlike VQA v1 and v2, each question in GQA and COVR only has one ground-truth answer. 

\begin{figure}[t]
	\centering  
	\includegraphics[width=0.9\columnwidth]{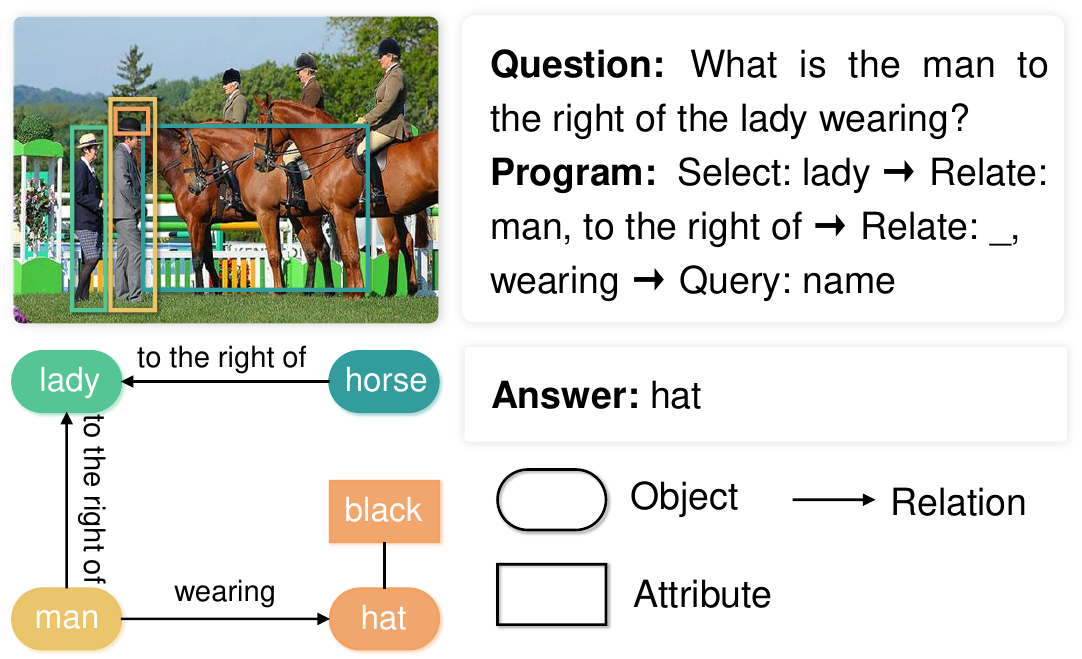}
	\caption{GQA example. This dataset provides a scene graph for each image and a functional program for each question. The program enumerates a series of logical operations (reasoning steps) necessary to obtain the answer.}
	\label{fig:gqa-eg}
\end{figure}

Fig. \ref{fig:gqa-eg} shows an example of GQA, which provides a scene graph \cite{thomee2016yfcc100m} for each image and a functional program for each question. The scene graph exhibits the relations between objects such as the ``to the right of" relation between ``man" and ``lady", while the program delineates a series of reasoning steps required to derive answers. In contrast, COVR increases the annotation of logical operators such as quantifiers and aggregations. Based on the above setting, VQA models can associate individual words from questions and answers with visual pointers, guiding more attention to pertinent regions within the image \cite{zhang2021vinvl}. Consequently, this dataset fosters the advancement of more explainable and robust models that necessitate diverse reasoning and multi-step inference abilities, specifically by enhancing the comprehension of both visual and linguistic components in a refined manner. Nevertheless, the lack of common sense would also increase the possibility of learning statistical regularity.

\noindent\textbf{CRIC}~\cite{gao2022cric}. To solve the above issue, this dataset\footnote{\url{https://cricvqa.github.io/}} was developed by generating questions and corresponding annotations from the scene graph of a given image and a knowledge graph respectively. In other words, this dataset not only includes identical components to GQA but also encompasses additional object-level knowledge triples such as the commonsense knowledge ``(subject: fork, relation: is used for, object: moving food)". Therefore, CRIC requires machines to ground common sense in the visual world and perform multi-hop reasoning on images and knowledge graphs.


\subsection{Out-of-Distribution Setting}
Various OOD datasets have been developed successively, where \emph{the distribution of the test split differs from or is even reversed to that of the training split}. These datasets aim to offer more challenging and intricate test scenarios from different perspectives, ranging from altering answer distributions to rephrasing questions. In this way, we can assess whether VQA methods can handle diverse real-world situations simultaneously or whether they are robust.

\noindent\textbf{VQA-CP v1 \& VQA-CP v2} \cite{agrawal2018don}. These two datasets were created by reorganizing VQA v1 and v2 according to the answer distribution, resulting in them sharing the same characteristics as VQA v1 and v2, such as multiple ground-truth answers for a single question. They are the first to explore the robustness evaluation of VQA methods, which motivates subsequent works \cite{lao2021superficial, liang0Z21, chocvpr,basucvpr} to focus on VQA robustness. VQA-CP v1\footnote{\url{http://data.lip6.fr/cadene/murel/vqacp2.tar.gz}} contains 370K questions with 205K images, while VQA-CP v2 contains 658K questions with 219K images.
\begin{figure}[t]
	\centering  
	\includegraphics[width=\columnwidth]{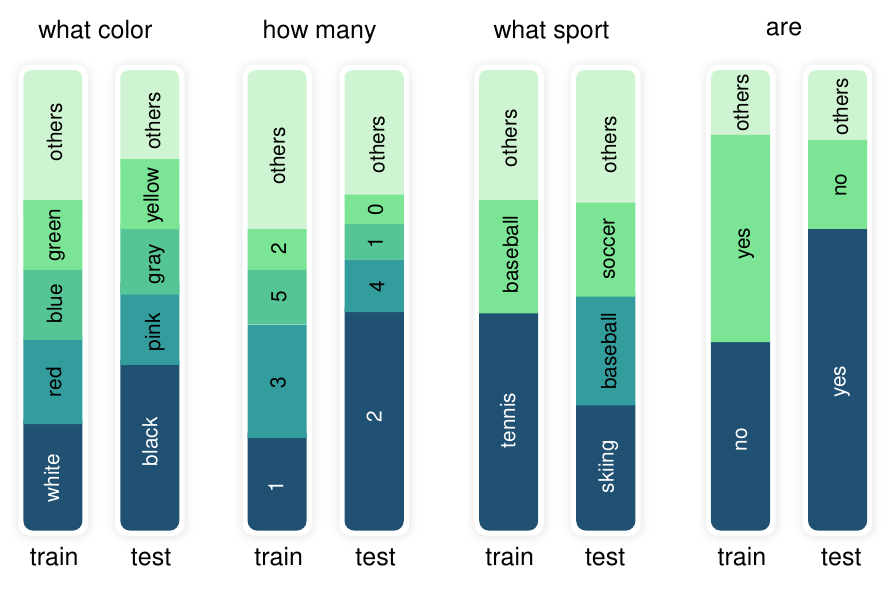}
	\caption{Answer distributions under specific types of questions in VQA-CP v1 training and test splits.}
	\label{fig:vqacp-eg}
\end{figure}

\fr{The distributions between the training and test splits of VQA-CP exhibit significant differences, and in some cases, they are even reversed. For instance, Fig. \ref{fig:vqacp-eg} shows that the most common answer to the question ``what sport" in the training split of VQA-CP v1 is ``tennis", while in the test split, ``skiing" emerges as the most prevalent answer. Therefore, these two datasets can be utilized to evaluate whether VQA methods predict answers through bias learning.}

However, these datasets are not without their limitations. Firstly, they consist of only two parts: a training set (67\%) and a test set (33\%), without including a validation set. This arrangement may lead to the tuning of hyperparameters on the test split. Secondly, the introduction of artificially crafted distribution shifts may not accurately reflect real-world scenarios. Lastly, models developed based on these datasets tend to be tailored toward their specific configurations due to the manual crafting of these shifts, potentially resulting in reduced generalization capabilities.

\begin{figure}[t]
	\centering  
	\includegraphics[width=\columnwidth]{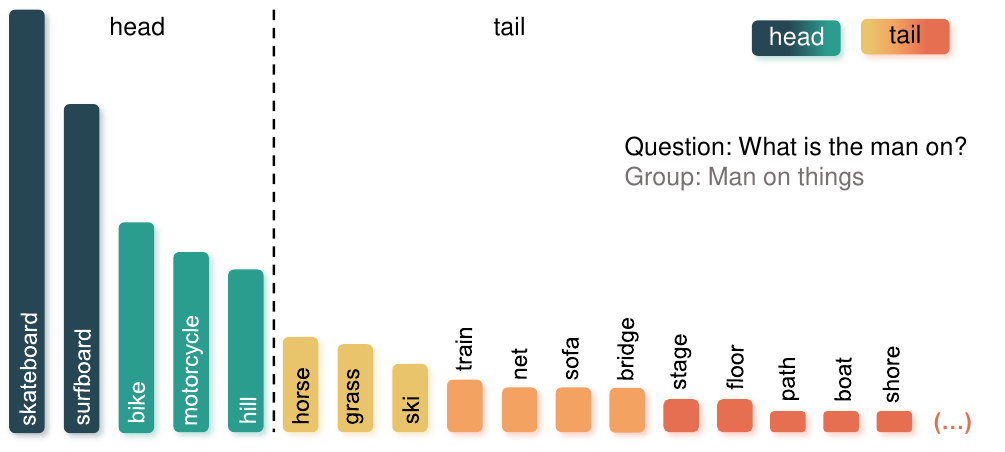}
	\caption{Answer distributions or frequencies within the ``Man on things" group. The ``tail" group is utilized to evaluate the OOD performance, while the ``head" group is employed to assess the ID performance.}
	\label{fig:gqaood-eg}
\end{figure}

\noindent\textbf{GQA-OOD}~\cite{kervadec2021roses}. To address the mentioned issues, this dataset\footnote{\url{https://github.com/gqa-ood/GQA-OOD}} was developed by reorganizing GQA \cite{hudson2019gqa} in a fine-grained manner such that distribution shifts are introduced in both the validation and test splits, and are tailored to different question groups. Specifically, GQA-OOD divides questions into two groups: ``head" and ``tail", according to the frequency of answers, instead of modifying the global distributions of the validation and test splits. \fr{ As illustrated in Fig. \ref{fig:gqaood-eg}, the answers to the ``man on things" question group are classified into two groups based on a criterion of $|a_i| \leq 1.2 \mu(a)$, where $|a_i|$ denotes the number of samples within the class $i$ such as ``skateboard" and ``surfboard", and $\mu(a)$ represents the average sample count for this group. }

GQA-OOD contains 53K questions with 9.7K images, which is split into two parts: validation (95\%) and testdev (5\%). Each part has two subsets: ``tail" and ``head". The ``tail" group is used to evaluate the OOD performance, while the ``head" group is employed to assess the ID performance. \fr{ This dataset requires models to be trained on the original GQA training split. In this way, it compels models to minimize bias in their test results while simultaneously exposing the models to bias captured within the training data \cite{kervadec2021roses}. Therefore, this dataset promotes the intrinsic development of debiasing methods rather than relying solely on the purification of training data. }

\begin{figure}[t]
	\centering  
	\includegraphics[width=\columnwidth]{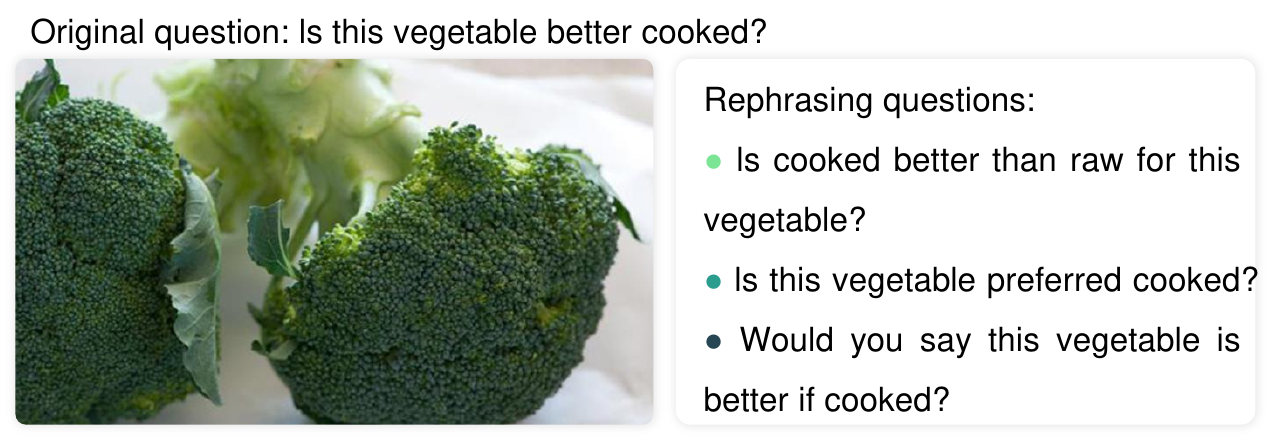}
	\caption{VQA-Rephrasings example. The original question is selected from the validation split of VQA v2, while the rephrased question is collected from three humans under the condition of preserving the original meaning.}
	\label{fig:vqare-eg}
\end{figure}

\noindent\textbf{VQA-Rephrasings}~\cite{shah2019cycle}. This dataset was introduced to assess the linguistic bias of VQA models. The term ``linguistic bias" refers to the phenomenon where the model's answer changes from correct to incorrect when the input question exhibits lexical variations \cite{shah2019cycle}. For example, given a picture containing a set of traffic lights, when the question changes from ``Is it safe to turn left?" to ``Can one safely turn left?", which essentially conveys the same meaning, the model's answer changes from ``yes" to ``no". This dataset was developed by first considering the validation split of VQA v2.0 as the base set, then sampling from this set to form a subset, and finally collecting three human-provided rephrasing for each question in this subset. \fr{ As illustrated in Fig. \ref{fig:vqare-eg}, the right questions are obtained by changing the expression of the left original question while retaining its original meaning. In this way, VQA-Rehprasings collects 162.0K questions with 40.5K images.} This dataset enables the evaluation of VQA methods for robustness and consistency across alternative rephrasing of questions with the same meaning.

\begin{figure}[t]
	\centering  
	\includegraphics[width=\columnwidth]{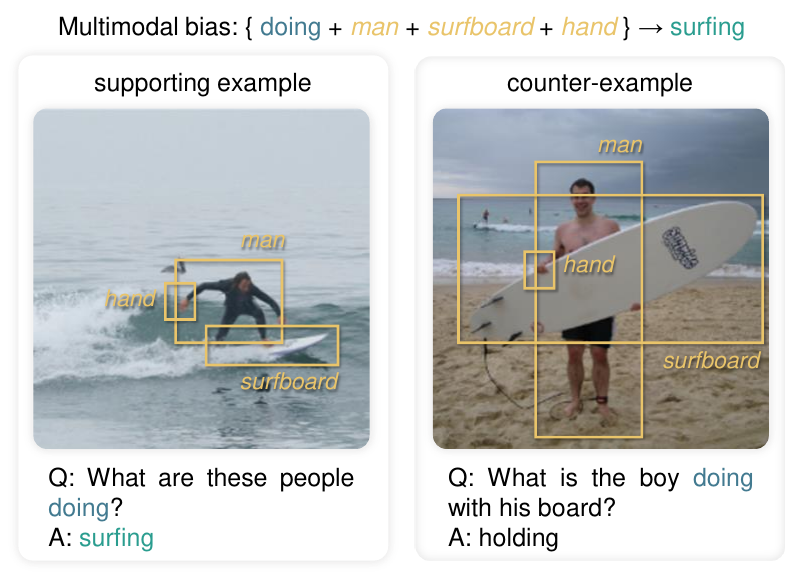}
    \caption{Example of a multimodal shortcut in VQACE, with its supporting examples and counter-examples. This shortcut is a combination of words ``doing" and objects including ``man", ``surfboard", and ``hand".}
	\label{fig:vqace-eg}
\end{figure}

\noindent\textbf{VQACE}~\cite{dancette2021beyond}. The previously mentioned datasets commonly overlook multimodal bias, instead prioritizing the assessment of language bias learning. \fr{ Multimodal bias is such a phenomenon that the frequent co-occurrence of textual and visual elements within the training data predicts specific answers accurately. Similar to language bias, multimodal bias often persists and transfers to the validation set, potentially impacting the generalizability of VQA models.} To address this issue, Dancette \textit{et al.} \cite{dancette2021beyond} proposed a method for identifying shortcuts, which predicts the correct answer based on the appearance of words in the question and visual elements in the image. Then, they built VQA counter-examples where the shortcut rules result in inaccurate answers as the evaluation protocol. \fr{ As depicted in Fig. \ref{fig:vqace-eg}, VQA models may learn or memorize certain words such as ``doing" and objects like ``man", ``surfboard", and ``hand" to predict answers. Although this can result in accurate predictions for some examples (as seen in the left example), it can also lead to incorrect responses (as observed in the right example). There is also an easy subset in which the correct answers can be derived through at least one of the shortcuts.} Statistics show that 90\% of the bias in this dataset\footnote{\url{https://github.com/cdancette/detect-shortcuts}} is multimodal, indicating that previous successful debiasing methods \cite{agrawal2018don,cadene2019rubi,niu2021introspective,gupta2022swapmix} in addressing the shift in language distributions such as the shift in VQA-CP and VQA-Rephrasings but may not be as effective in reducing natural shortcuts from VQA.

\noindent\textbf{VQA-VS}~\cite{si2022language}. Like VQACE, this dataset\footnote{\url{https://github.com/PhoebusSi/VQA-VS}} takes into account the natural shortcuts in VQA, such as language, vision, and multimodality bias. It goes a step further by introducing several concrete shortcuts for each bias, such as the question-type shortcut in language bias and the key-object-pair shortcut in vision bias. \fr{For instance, the multimodal bias of VQACE in Fig. \ref{fig:vqace-eg} belongs to the shortcut of ``composite of keyword and key object" in this dataset.} These shortcuts may be used to assess OOD performance in a more refined manner. In addition, this dataset also includes a split to evaluate ID performance.

\fr{However, the datasets mentioned above have certain limitations \cite{li2021adversarial}. Firstly, these datasets are mainly developed based on heuristic rules \cite{agarwal2020towards, agrawal2018don,gokhale2020vqa}. Furthermore, synthetic images or questions are often used in these datasets \cite{agarwal2020towards,gokhale2020vqa,hudson2019gqa,johnson2017clevr}, instead of being generated by humans. They may not represent real-world scenarios accurately.}

\begin{figure}[t]
	\centering  
	\includegraphics[width=\columnwidth]{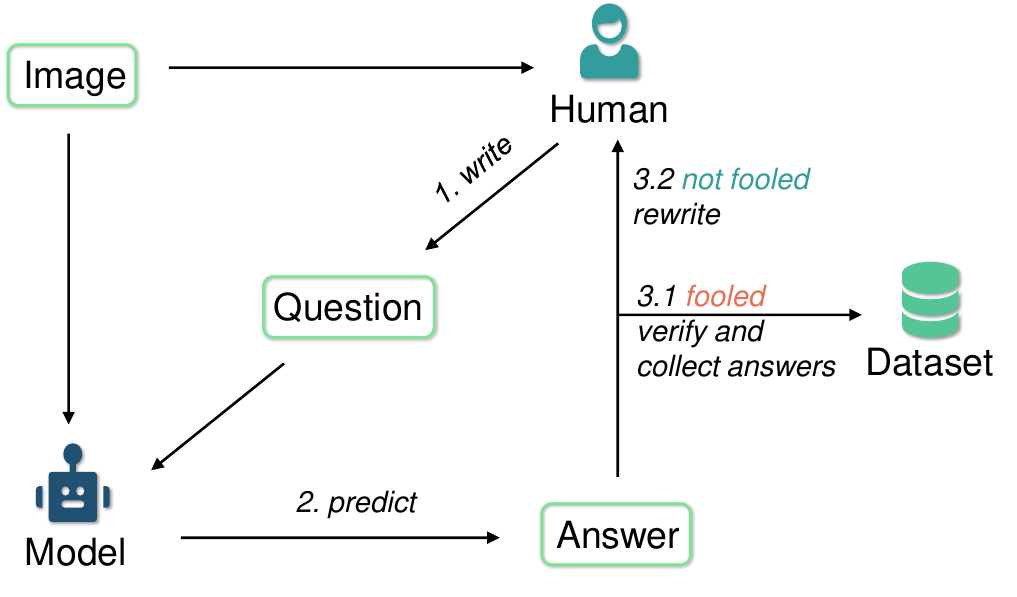}
	\caption{Illustration of the human-and-model-in-the-loop procedure. ``fooled" can be regarded as a successful attack.}
	\label{fig:hil}
\end{figure}

\noindent\textbf{AVQA}~\cite{li2021adversarial}. To address the above issues, this dataset\footnote{\url{https://adversarialvqa.org/}} was gathered iteratively using an adversarial ``human-and-model-in-the-loop" procedure that is shown in Fig. \ref{fig:hil}. The process of data collection can be viewed as a game played in $n$ rounds between two parties: a human annotator and a well-trained model. AVQA consists of 243.0K questions accompanied by 37.9K images and is developed through three rounds. Specifically, the image-question pairs are gathered once VQA models that are trained on VQA v2 \cite{goyal2017making} and Visual Genome \cite{krishna2017visual} cannot predict answers accurately. In other words, this image-question pair can be regarded as a successful attack. \fr{To avoid over-fitting, the VQA model is randomly chosen from LXMERT \cite{tan2019lxmert}, UNITER-B \cite{chen2020uniter}, and UNITER-L \cite{chen2020uniter}. To collect harder questions, the model is re-trained based on the data collected from previous rounds. Note that the question is written by humans rather than rule-based questions such as in GQA. Therefore, AVQA may be more challenging than previous datasets \cite{malinowski2014multi,gao2015you,zhu2016visual7w,goyal2017making,krishna2017visual,johnson2017clevr}. }
\begin{table*}[t]
  \centering    
  \caption{Comparison of datasets and evaluation metrics. ID denotes in-distribution, and OOD represents out-of-distribution. A single graph denotes a scene graph of an image. Double graphs denote scene and knowledge graphs. CC \protect\cite{sharma2018conceptual} is the dataset of conceptual captions. Fakeddit \protect\cite{nakamuraLW20} is a multimodal dataset for fake news detection. Natural shortcuts include language, vision, and multimodality bias. MPT denotes the mean-per-type accuracy.}
  \label{tab:datasets}%
  \resizebox{\textwidth}{!}{
  \begin{tabular}{rcccccccc}
    \toprule
    \textbf{Dataset}                          & \textbf{Development}  & \textbf{Image Source}    & \textbf{Images} & \textbf{Questions} & \textbf{Focus} & \textbf{ID}    & \textbf{OOD}  & \textbf{Metrics}\\
    \midrule
    VQA v1 \cite{antol2015vqa}       & asking human questions       & COCO            & 204.7K & 614.2K & vision     & \Checkmark     & \XSolidBrush & open-ended accuracy  \\
    VQA v2 \cite{goyal2017making}    & complement on VQA v1         & COCO            & 479K   & 1.1M   & vision        & \Checkmark     & \XSolidBrush & open-ended accuracy \\
    TDIUC \cite{kafle2017analysis}   & reorganization on several VQA datasets& COCO and Visual Genome   & 167K   & 1.6M   & vision        & \Checkmark     & \XSolidBrush & arithmetic MPT \\
    GQA \cite{hudson2019gqa}         & generate questions from a single graph & Visual Genome  & 113K   & 22M    & composition   & \Checkmark & \XSolidBrush & composite metrics \\
    COVR \cite{bogin2021covr}        & generate questions from a subgraph & Visual Genome   & 88.5K    & 262K & composition & \Checkmark     & \XSolidBrush & standard accuracy \\
    CRIC \cite{gao2022cric}          & generate questions from double graphs & Visual Genome   & 96K    & 494.3K & commonsense & \Checkmark     & \XSolidBrush & composite metrics \\
    VQA-CP v1 \cite{agrawal2018don}  & reorganization on VQA v1     & COCO            & 205K   & 370K   & language bias & \XSolidBrush   & \Checkmark   & open-ended accuracy \\
    VQA-CP v2 \cite{agrawal2018don}  & reorganization on VQA v2     & COCO            & 219K   & 658K   & language bias & \XSolidBrush   & \Checkmark   & open-ended accuracy \\
    GQA-OOD \cite{kervadec2021roses} & reorganization on GQA        & Visual Genome   & 82.2K  & 996.8K & language bias & \Checkmark     & \Checkmark   & composite metrics \\
    VQA-Rephrasings \cite{shah2019cycle}& rephrasing on VQA v2      & COCO            & 40.5K  & 162.0K & language bias & \XSolidBrush   & \Checkmark   & consensus score        \\
    VQA-CE \cite{dancette2021beyond} & reorganization on VQA v2     & COCO            & -      & -      & natural shortcuts & \XSolidBrush   & \Checkmark & open-ended accuracy \\
    VQA-VS \cite{si2022language}     & reorganization on VQA v2     & COCO            & -      & -      & natural shortcuts & \Checkmark     & \Checkmark & open-ended accuracy \\
    AVQA \cite{li2021adversarial}    & human-and-model-in-the-loop  & CC/Fakeddit/VCR & 37.9K  & 243.0K & adversarial robustness     & \XSolidBrush     & \Checkmark      & open-ended accuracy \\
    AdVQA \cite{sheng2021human}      & human-and-model-in-the-loop  & COCO            & 41.8K  & 46.8K  & adversarial robustness     & \XSolidBrush   & \Checkmark        & open-ended accuracy \\
    \bottomrule
    \end{tabular}%
  }
\end{table*}

\noindent\textbf{AdVQA}~\cite{sheng2021human}. Similar to AVQA, this dataset\footnote{\url{https://adversarialvqa.org/}} was also developed by the human-and-model-in-the-loop procedure \cite{kiela2021dynabench}. In other words, both of them are developed from the 
perspective of adversarial robustness. This dataset contains 46.8K questions accompanied by 41.8K images. In comparison, the developing procedure for this dataset has an additional question validation phase, which manually determines whether the image is necessary and sufficient to answer the question. 

In a nutshell, each dataset has a distinct focus and can be used to assess the ability of methods from a particular angle.
For example, in ID settings, TDIUC \cite{kafle2017analysis} presents the challenge in task-driven reasoning for VQA; GQA \cite{hudson2019gqa} and COVR \cite{bogin2021covr} contribute to real-world visual reasoning and compositional question answering; CRIC \cite{gao2022cric} necessitates grounding common sense in the visual domain and performing multi-hop reasoning on images and knowledge graphs. Similarly, in out-of-distribution settings, VQA-CP v1 and VQA-CP v2 introduce distribution shifts in training and testing splits; GQA-OOD \cite{kervadec2021roses} compels models to minimize biases while being exposed to biases present in the training data one step further; VQA-Rephrasings \cite{shah2019cycle} focuses on emphasizing robustness and consistency in handling varied expressions of the same semantic content; VQACE \cite{dancette2021beyond} and VQA-VS \cite{si2022language} challenge models to mitigate language, vision, and multimodal biases effectively; AVQA \cite{li2021adversarial} and AdVQA \cite{sheng2021human} present a more challenging set of questions generated iteratively using an adversarial ``human-and-model-in-the-loop" procedure. More details of datasets, such as the source of images and the focus, are shown in Table \ref{tab:datasets}.

\section{Evaluations}
\label{sec:eval}
The evaluation is usually associated with the annotation of datasets, which ranges from open-ended accuracy to composite metrics. Existing debiasing works \cite{liang2020learning,chen2020counterfactual,niu2021introspective,lao2022vqa} typically employ a combination of ID and OOD datasets to assess robustness. As a result, a trade-off metric such as the harmonic mean is used to assess the performance of methods comprehensively. The evaluation metrics of the mentioned datasets in Section \ref{sec:data} are shown in Table \ref{tab:datasets}.

\noindent\textbf{Open-Ended Accuracy.} Given an image and its related question, humans may provide various answers in a real-world scenario. Inspired by this, to comprehensively evaluate VQA methods, the questions in VQA v1 \cite{antol2015vqa} and v2 \cite{goyal2017making} are annotated by ten humans, resulting in the employment of open-ended accuracy. The main design intention of this metric is that a model will obtain higher accuracy if its predicted answers have a greater consensus. For example, ``off duty" and ``coming" are provided once and three times, respectively, for the question that is on the right of Fig. \ref{fig:vqa-v1-eg} by different annotators. If the mentioned answers are predicted by different models respectively, the model with ``coming" will obtain the higher accuracy. Furthermore, various datasets, such as VQA-CP \cite{agrawal2018don} and VQA-CE \cite{dancette2021beyond}, also use this metric because their development is related to VQA v1 and v2 datasets. The open-ended accuracy is computed as follows:
\begin{equation}
	\text{open-ended accuracy} = \min \left\{ \frac{n_{\mathrm{a}}}{3}, 1 \right\},
\end{equation}
where $n_{\mathrm{a}}$ denotes the number of predicted answers that are identical to human-provided answers for questions and 3 is the minimum number of consensuses.

\noindent\textbf{Composite Metrics.} Unlike the datasets mentioned above, GQA \cite{hudson2019gqa} and COVR \cite{bogin2021covr} only contain an answer for each question, making it possible to use the ``standard accuracy" as an evaluation metric. In addition, GQA introduces five new metrics to gain a better understanding of visual reasoning techniques and highlights the functions that coherent reasoning methods should have.
\begin{itemize}
    \item ``Consistency" assesses the consistency of predicted answers across different questions. For example, given a question-answer pair ``(Is there a banana to the right of the white cup?, Yes)" and an image, we can infer the answers to questions such as ``Is there a fruit to the right of the white cup?" and ``Is the white cup to the left of the banana?". 
    \item ``Validity" and ``Plausibility" are used to evaluate whether the predicted answer is reasonable enough. The former checks whether the predicted answer is within the predefined answer scope of the questions, such as by providing colors to a color-type question. The latter goes a step further in determining whether the answer makes sense. It checks whether the model prediction occurs at least once with the object over the whole dataset scene graphs; for example, objects like ``pandas" do not have the ability or attribute such as ``talk" in the scene graph.
    \item ``Distribution" assesses the overall match between the true answer distribution and the method's predicted distribution using Chi-Square statistics \cite{lancaster2005chi}. This metric enables us to evaluate the ability of models to predict not only the most frequent answers but also the less common ones.
    \item ``Grounding" evaluates how well the model concentrates on regions of the image that are crucial to the question. This metric is calculated by summing the attention score across the region and then averaging them across all questions in the dataset.
\end{itemize}

As introduced in the above subsection, GQA-OOD is developed by re-splitting the validation and test splits of GQA into \emph{head} and \emph{tail} groups respectively. Therefore, we can also use these five metrics to evaluate VQA methods besides the ``head" and ``tail" accuracy. 

In addition to these metrics, there also exist other composite metrics to evaluate VQA performance. Specifically, CRIC \cite{gao2022cric} requires methods not only to predict answers but also to provide intermediate grounding results to mitigate the impact of commonsense prior and enables us to fairly evaluate whether methods truly understand the image and commonsense. In other words, a VQA system should provide two outputs for each question including an answer and a chosen object from the image. As a result, a question is considered to be correctly answered when the two predictions are both correct.

\noindent\textbf{Arithmetic MPT.} TDIUC \cite{kafle2017analysis} explicitly categorizes questions into 12 distinct categories, providing us with a comprehensive protocol to evaluate VQA methods from various perspectives. To compensate for the imbalanced distribution of question types, the standard accuracy is calculated separately for each category. Furthermore, an arithmetic mean-per-type accuracy is computed to derive a unified accuracy metric.

\noindent\textbf{Consensus Score.} To measure the robustness of methods across various question rephrasing, VQA-Rephrasings \cite{shah2019cycle} introduces a metric dubbed ``consensus score" based on the premise that the answer to all rephrasing of the same question should be the same. \fr{For instance, a system should provide the same answer given the four questions in Fig. \ref{fig:vqare-eg}.} This consensus score $cs$ is defined as the ratio of the number of subsets where all the answers are correct and the total number of subsets with size $k$:
\begin{equation}
    \begin{split}
    & cs(k) = \sum_{\mathcal{Q}^{'}\subset\mathcal{Q},\left|\mathcal{Q}^{'}\right|=k} \frac{s(\mathcal{Q}^{'})}{^nC_k}, \\
    & \text{with } s(\mathcal{Q}') = \left\{
                \begin{array}{ll}
                  1 & \text{if} \quad \forall q\in\mathcal{Q}^{'} \quad \phi(q)>0\text{,} \\
                  0 & \text{otherwise.}
                \end{array}\right.,
    \end{split}
\end{equation}
where $^nC_k$ denotes the number of subsets with size $k$ sampled from a set with size $n$, $\mathcal{Q}^{'}$ denotes a group of questions contained in $\mathcal{Q}$ that consists of $n$ rephrasings, and $\phi(q)$ is the open-ended accuracy.

\section{Debiasing Methods}
\label{sec:meth}
In recent years, based on generic methods such as UpDn \cite{anderson2018bottom}, BAN \cite{kim2018bilinear}, SMRL \cite{cadene2019murel}, and LXMERT \cite{tan2019lxmert}, a variety of VQA debiasing methods \cite{song2023recovering,jiang2021x,Qraitem2023} have been proposed to improve robustness. We divide these methods into four categories according to the debiasing technique: \emph{ensemble learning, data augmentation, self-supervised contrastive learning, and answer re-ranking}. Table \ref{tab:methods} presents their typology as well as their performance on both the VQA-CP v2 \cite{agrawal2018don} test split and the VQA v2 \cite{goyal2017making} validation split. Due to the limited exploration of existing studies on the OOD dataset including VQACE\cite{dancette2021beyond}, AdVQA\cite{sheng2021human}, VQA-VS\cite{si2022language}, GQA-OOD\cite{kervadec2021roses}, AVQA\cite{li2021adversarial} and VQA-Rephrasings\cite{shah2019cycle}, we encourage future explorations and research endeavors in tackling the intricate challenges they pose. Table \ref{tab:results on other datasets} presents the performance of some methods on these datasets.
\begin{table*}[!ht]
  \centering
  \caption{The typology and performance of existing debiasing methods. HM denotes the harmonic mean of overall accuracy. \CIRCLE~represents the main category that a method belongs to, while \Circle~ denotes the method that also uses the other debiasing technique. The result marked in bold is the best performance on the dataset. The result of the method with two references following is reported by the latter reference. LM denotes the Learned Mixin \cite{hinton2002training}, while LMH denotes LM with an entropy penalty.} \label{tab:methods}%
  \resizebox{\textwidth}{!}{
    \begin{tabular}{rcccccccccc}
    \toprule
    \textbf{Method} & \textbf{Base} & \textbf{Year}   & \makecell[c]{\textbf{Ensemble} \\ \textbf{Learning}} & \makecell[c]{\textbf{Data} \\ \textbf{Augmentation}} &\makecell[c]{ \textbf{Self-Supervised} \\ \textbf{Contrastive Learning}}  &\makecell[c]{\textbf{Answer} \\ \textbf{Re-Ranking}}  
    & \makecell[c]{\textbf{VQA-CP v2} \\ \textbf{test}}  & \makecell[c]{\textbf{VQA v2} \\ \textbf{validation}}  & \textbf{HM} \\
    \midrule
    Template-based \cite{kafle2017data},\cite{kil2021discovering}   & UpDn    & 2017    &         & \CIRCLE &         &         & 39.75  & 63.83  & 48.99  \\
    GVQA \cite{agrawal2018don}                                      & SAN     & 2018    &         &         &         & \CIRCLE & 31.30  & 48.24  & 37.97 \\
    AdvReg \cite{ramakrishnan2018},\cite{wu2019self}                & UpDn    & 2018    & \CIRCLE &         &         &         & 41.17  & 62.75  & 49.72 \\
    AttAlign \cite{selvaraju2019taking}                             & UpDn    & 2019    &         &         &         & \CIRCLE & 39.37  & 63.24  & 48.53 \\
	RUBi \cite{cadene2019rubi}                                      & UpDn    & 2019    & \CIRCLE &         &         &         & 47.11  & 61.16  & 53.22 \\
    HINT \cite{selvaraju2019taking}                                 & UpDn    & 2019    &         &         &         & \CIRCLE & 46.73  & 63.38  & 53.80 \\
    LMH \cite{clark2019don}                                         & UpDn    & 2019    & \CIRCLE &         &         &         & 52.87  & 55.99  & 54.39 \\
    SCR \cite{wu2019self}                                           & UpDn    & 2019    &         &         &         & \CIRCLE & 49.45  & 62.20  & 55.10 \\
    ASL \cite{teney2019actively}                                    & UpDn    & 2019    &         & \CIRCLE &         &         & 46.00  & -      & - \\   \midrule[0.2pt]
    DLR \cite{jing2020overcoming}                                   & SAN     & 2020    &         &         &         & \CIRCLE & 34.83  & 49.27  & 40.81 \\
    CSS+CL \cite{liang2020learning}                                 & UpDn    & 2020    & \Circle & \Circle & \CIRCLE &         & 40.49  & -      & -  \\
    CVL \cite{abbasnejad2020counterfactual}                         & UpDn    & 2020    &         & \CIRCLE &         &         & 42.12  & -      & - \\
    CVL \cite{abbasnejad2020counterfactual}                         & Pythia  & 2020    &         & \CIRCLE &         &         & -      & 68.77  & - \\
    GradSup \cite{teney2020learning}                                & UpDn    & 2020    &         & \CIRCLE &         &         & 46.80  & -      & - \\
    RankVQA \cite{qiao2020rankvqa}                                  & UpDn    & 2020    &         &         &         & \CIRCLE & 43.05  & 65.42  & 51.93  \\
    DLR \cite{jing2020overcoming}                                   & UpDn    & 2020    &         &         &         & \CIRCLE & 48.87  & 57.96  & 53.03 \\
    SimpleReg \cite{shrestha2020negative}                           & UpDn    & 2020    &         &         &         & \CIRCLE & 48.90  & 62.60  & 54.91  \\
    RMFE \cite{gat2020removing}                                      & LMH     & 2020    & \Circle &         &         & \CIRCLE & 58.21  & 53.15  & 55.57 \\
    VGQE \cite{kv2020reducing}                                      & SMRL    & 2020    &         & \CIRCLE &         &         & 50.11  & 63.18  & 55.89 \\
    RandImg \cite{teney2020value}                                   & UpDn    & 2020    &         & \CIRCLE &         &         & 55.37  & 57.24  & 56.29  \\
    CSS+CL \cite{liang2020learning}                                 & LMH     & 2020    & \Circle & \Circle & \CIRCLE &         & 59.18  & 57.29  & 58.22  \\
    VILLA \cite{gan2020large}                                       & -       & 2020    &         & \CIRCLE &         &         & 49.10  & 74.69  & 59.25  \\
    CSS \cite{chen2020counterfactual}                               & LMH     & 2020    & \Circle & \CIRCLE &         &         & 58.95  & 59.91  & 59.43  \\
    MANGO \cite{li2020closer}                                       & -       & 2020    &         & \CIRCLE &         &         & 52.76  & 74.26  & 61.69  \\
    MUTANT \cite{gokhale2020mutant}                                 & UpDn    & 2020    &         & \CIRCLE &         &         & 61.72  & 62.56  & 62.14 \\ \midrule[0.2pt]
    X-GGM \cite{jiang2021x}                                         & -       & 2021    &         & \CIRCLE &         &         & -      & -      & - \\
    MCE \cite{clark2020learning}                                    & -       & 2021    & \CIRCLE &         &         &         & -      & -      & - \\
    LPF \cite{liang0Z21}                                            & BAN     & 2021    & \Circle &         &         & \CIRCLE & 50.76  & -      & -  \\
    LPF \cite{liang0Z21}                                            & SMRL    & 2021    & \Circle &         &         & \CIRCLE & 53.38  & -      & -  \\
    LBCL \cite{lao2021superficial}                                  & -       & 2021    & \CIRCLE &         &         &         & 60.74  & -      & -  \\
    AdVQA \cite{GuoNCJZB21},\cite{basu2023rmlvqa}                   & -       & 2021    & \CIRCLE &         &         &         & 54.67  & 46.98  & 50.53 \\
    Unshuffling \cite{teney2021unshuffling}                         & UpDn    & 2021    &         & \CIRCLE &         &         & 43.37  & 63.47  & 51.53  \\
    IntroD \cite{niu2021introspective}                              & RUBi    & 2021    & \CIRCLE &         &         &         & 48.54  & 61.86  & 54.40  \\
    IntroD \cite{niu2021introspective}                              & LMH     & 2021    & \CIRCLE &         &         &         & 51.31  & 62.05  & 56.17  \\
    LPF \cite{liang0Z21}                                            & UpDn    & 2021    & \Circle &         &         & \CIRCLE & 51.57  & 62.63  & 56.56  \\
    CF \cite{niu2021counterfactual}                                 & SMRL    & 2021    & \CIRCLE &         &         &         & 53.55  & 60.94  & 57.01  \\
    CIKD \cite{pan2021distilling}                                   & UpDn    & 2021    & \CIRCLE &         &         &         & 54.05  & 61.29  & 57.44  \\
    SimpleAug \cite{kil2021discovering}                             & LMH     & 2021    &         & \CIRCLE &         &         & 53.70  & 62.63  & 57.82  \\
    SimpleAug \cite{kil2021discovering}                             & UpDn    & 2021    &         & \CIRCLE &         &         & 52.65  & 64.34  & 57.91  \\
    GGE \cite{han2021greedy}                                        & UpDn    & 2021    & \CIRCLE &         &         &         & 57.32  & 59.11  & 58.20  \\
    CF \cite{niu2021counterfactual}                                 & UpDn    & 2021    & \CIRCLE &         &         &         & 55.05  & 63.73  & 59.07  \\
    CCB \cite{yang2021learning}                                     & HINT    & 2021    & \Circle &         &         & \CIRCLE & 59.12  & 59.17  & 59.14  \\
    CCB \cite{yang2021learning}                                     & UpDn    & 2021    & \Circle &         &         & \CIRCLE & 57.99  & 60.73  & 59.32  \\
    LP-Focal \cite{lao2021language}                                 & UpDn    & 2021    &         &         &         & \CIRCLE & 58.45  & 62.45  & 60.38  \\
    SSL \cite{zhu2021overcoming}                                    & UpDn    & 2021    &         & \Circle & \CIRCLE &         & 57.59  & 63.73  & 60.50 \\
    IntroD \cite{niu2021introspective}                              & CSS     & 2021    & \CIRCLE &         &         &         & 60.17  & 62.57  & 61.35  \\
    D-VQA \cite{wenXTWW21}                                          & UpDn    & 2021    & \CIRCLE &         & \Circle &         & 61.91  & 64.96  & 63.40  \\
    SAR \cite{si2021check}                                          & LMH     & 2021    & \Circle &         &         & \CIRCLE & 66.73  & 69.22  & 67.95  \\
    SimpleAug \cite{kil2021discovering}                             & LXMERT  & 2021    &         & \CIRCLE &         &         & 62.24  & \textbf{74.98}  & 68.02  \\ \midrule[0.2pt]
    SwapMix \cite{gupta2022swapmix}                                 & -       & 2022    &         & \CIRCLE &         &         & -      & -      & - \\
    Loss-Rescaling \cite{guo2022loss}                               & UpDn    & 2022    &         &         &         & \CIRCLE & 47.09  & 55.50  & 50.95  \\
    Loss-Rescaling \cite{guo2022loss}                               & LMH     & 2022    &         &         &         & \CIRCLE & 53.26  & 56.81  & 54.98  \\
    Loss-Rescaling \cite{guo2022loss}                               & CSS     & 2022    &         &         &         & \CIRCLE & 50.73  & 61.14  & 55.45  \\
    Loss-Rescaling \cite{guo2022loss}                               & LM      & 2022    &         &         &         & \CIRCLE & 53.17  & 59.45  & 56.13  \\
    Loss-Rescaling \cite{guo2022loss}                               & CSS+LMH & 2022    &         &         &         & \CIRCLE & 56.55  & 55.96  & 56.25  \\
    ECD \cite{kolling2022efficient}                                 & UpDn+LMH& 2022    & \Circle & \CIRCLE &         &         & 59.92  & 57.38  & 58.62  \\
    MMBS \cite{si2022towards}                                       & LMH     & 2022    & \Circle & \Circle & \CIRCLE &         & 56.44  & 61.87  & 59.03  \\
    KDDAug \cite{chen2022rethinking}                                & RUBi    & 2022    &         & \CIRCLE &         &         & 59.25  & 60.25  & 59.75 \\
    KDDAug \cite{chen2022rethinking}                                & LMH     & 2022    &         & \CIRCLE &         &         & 59.54  & 62.09  & 60.79 \\
    VQA-BC \cite{lao2022vqa}                                        & LMH     & 2022    & \Circle & \Circle & \CIRCLE &         & 60.81  & 61.74  & 61.27  \\
    AttReg \cite{liu2022answer}                                     & LMH     & 2022    & \Circle & \CIRCLE &         &         & 60.00  & 62.74  & 61.34  \\
    KDDAug \cite{chen2022rethinking}                                & UpDn    & 2022    &         & \CIRCLE &         &         & 60.24  & 62.86  & 61.52 \\
    KDDAug \cite{chen2022rethinking}                                & CSS     & 2022    &         & \CIRCLE &         &         & 61.14  & 62.17  & 61.65 \\
    Loss-Rescaling \cite{guo2022loss}                               & LXMERT  & 2022    &         &         &         & \CIRCLE & 66.40  & 69.76  & \textbf{68.04}  \\ 
    RMLVQA \cite{basu2023rmlvqa}                                    & UpDn    & 2023    & \Circle &         &         & \CIRCLE & 60.41  & 59.99  & 60.20  \\
    GGD \cite{han2023general}                                       & UpDn    & 2023    & \CIRCLE &         &         &         & 59.37  & 62.15  & 60.73  \\
    GenB \cite{cho2023generative}                                   & LXMERT  & 2023    & \CIRCLE &         &         &         & \textbf{71.16}  & -      & -  \\
    LSP \cite{liu2023flexible}                                      & UpDn    & 2023    & \Circle & \Circle &   \CIRCLE &       & 61.95  & 65.26  & 63.56  \\
    \bottomrule
    \end{tabular}%
  }
\end{table*}

\begin{table*}
\caption{The performance of some methods on the test splits of VQACE\cite{dancette2021beyond}, AdVQA\cite{sheng2021human}, VQA-VS\cite{si2022language}, GQA-OOD\cite{kervadec2021roses}, AVQA\cite{li2021adversarial} and VQA-Rephrasings\cite{shah2019cycle}. The result marked in bold is the best performance on the dataset. }
\label{tab:results on other datasets}
\centering
 \resizebox{\textwidth}{!}{
\begin{tabular}{crccrccrccrc}
\toprule
\textbf{Dataset} &\textbf{Methods} &\textbf{Accuracy} &\textbf{Dataset} &\textbf{Methods} &\textbf{Accuracy} &\textbf{Dataset} &\textbf{Methods} &\textbf{Accuracy} &\textbf{Dataset} &\textbf{Methods} &\textbf{Accuracy} \\ \midrule
\multirow{10}{*}{VQA-CE} & VilBERT \cite{lu2019vilbert} & \textbf{67.77} & \multirow{10}{*}{AdVQA} & VisualBERT \cite{li2019visualbert} & \textbf{27.36} & \multirow{6}{*}{VQA-VS} & LXMERT \cite{tan2019lxmert} & \textbf{53.70} & \multirow{6}{*}{AVQA} & VILLA\textsubscript{Base} \cite{gan2020large} & \textbf{26.08} \\
  & LfF \cite{nam2020learning} & 63.57 &             & VilBERT \cite{lu2019vilbert} & 27.36 &                      & BAN \cite{kim2018bilinear} & 48.53 &    & UNITER\textsubscript{Large} \cite{chen2020uniter} & 24.78 \\
  & UpDn \cite{anderson2018bottom} & 63.52 &         & ViLT \cite{kim2021vilt} & 27.11 &                           & UpDn \cite{anderson2018bottom} & 46.80 & & ClipBERT \cite{lei2021less} & 24.35 \\
  & RandImg \cite{teney2020value} & 63.34 &          & UNITER\textsubscript{Large} \cite{chen2020uniter} & 26.94 & & LPF \cite{liang0Z21} & 45.85 &          & LXMERT \cite{tan2019lxmert} & 24.13 \\
  & SimpleReg \cite{shrestha2020negative} & 62.96 &  & BERT \cite{kenton2019bert} & 26.90  &                        & LMH \cite{clark2019don} & 45.85 &       & UNITER\textsubscript{Base} \cite{chen2020uniter} & 24.10 \\
  & RUBi \cite{cadene2019rubi} & 61.88 &             & MMBT \cite{kiela2019supervised} & 26.70 &                    & SSL \cite{zhu2021overcoming} & 45.62 &  & UpDn \cite{anderson2018bottom} & 22.78 \\ 
  \cmidrule(lr){7-9} \cmidrule(l){10-12}
  & LMH \cite{clark2019don} & 61.15 &   & MCAN \cite{yu2019deep} & 26.64 & \multirow{4}{*}{GQA-OOD} & LXMERT \cite{tan2019lxmert} & \textbf{54.60} & \multirow{4}{*}{VQA-Rephrasings} & BAN \cite{kim2018bilinear} & \textbf{56.59} \\
  & RMFE \cite{gat2020removing} & 60.96 &        & VILLA\textsubscript{Large} \cite{gan2020large} & 25.79 &    & MCAN \cite{yu2019deep} & 50.80  &         & UpDn \cite{anderson2018bottom} & 52.58 \\
  & SAN \cite{yang2016stacked}  & 55.61 &        & UNITER\textsubscript{Base} \cite{chen2020uniter} & 25.16 &  & BAN \cite{kim2018bilinear} & 50.20 &      & MUTAN \cite{ben2017mutan} & 46.87 \\
  & CSS \cite{chen2020counterfactual} & 53.55 &  & VILLA\textsubscript{Base} \cite{gan2020large}  & 25.14 &    & UpDn \cite{anderson2018bottom} & 46.40 &  &          &       \\ \bottomrule
\end{tabular}
}
\end{table*}

\subsection{Ensemble Learning}
\begin{figure}[t]
    \centering
	\includegraphics[width=\columnwidth]{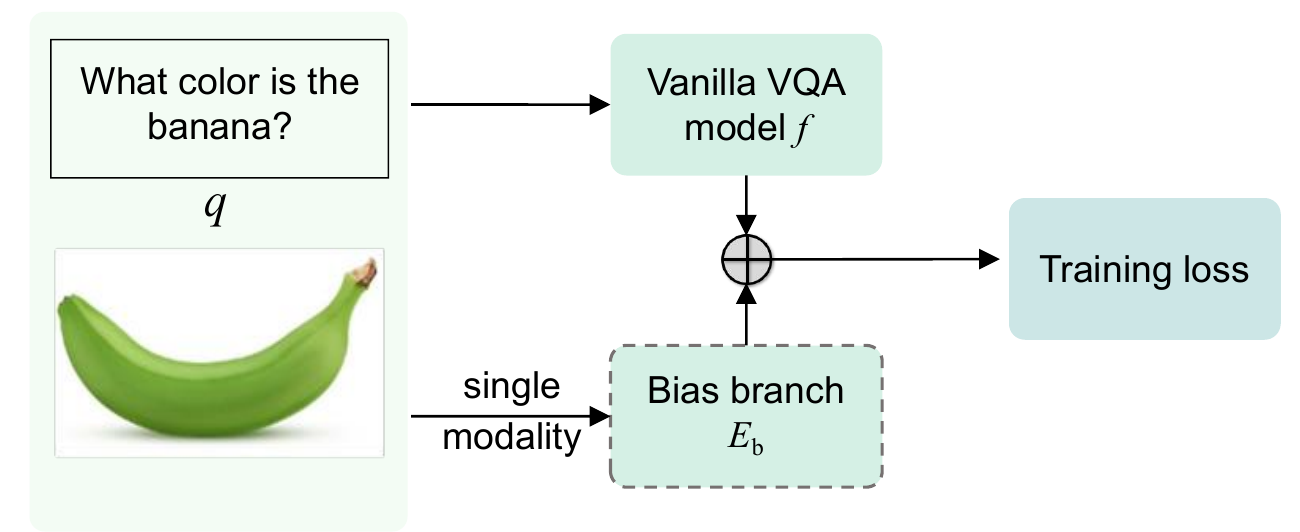}
	\caption{Illustration of ensemble learning-based debiasing methods. The bias branch takes single modalities as inputs to capture bias learning in the training stage. In the test stage, the branch is removed, and answers are provided by the vanilla VQA model. $\oplus$ denotes the combination operation such as a sigmoid function.}
	\label{fig:EB}
\end{figure}
Ensemble learning-based methods are the first to investigate the robustness of VQA. They typically apply the combination $\Phi$ of a bias branch $E_{\mathrm{b}}$ and a vanilla VQA method $f$ defined in Equation (\ref{eq:gp}) to predict answers $\hat{a}_i$ comprehensively. $E_{\mathrm{b}}$ is used to capture the bias learning, while $f$ is applied to perform normal question answering. This process can be denoted as follows:
\begin{equation}
    \hat{a}_i = \Phi \left( E_{\mathrm{b}} \left(E_{\mathrm{v}}(v_i), E_{\mathrm{q}}(q_i) \right), f(v_i, q_i) \right) . \label{eq:ensemble}
\end{equation}

The illustration of the ensemble learning-based method is shown in Fig. \ref{fig:EB}. \emph{In the training stage}, a single modality, such as questions alone, is fed into the bias branch, irrespective of images. \fr{In this way, the accurate prediction of the answer can be achieved to a great extent without relying on the visual information provided by the image modality.} In other words, the bias or statistical correlations in the training data are captured. Then, the bias branch and the vanilla method are trained in an ensemble manner. For example, their predictions are combined by a sigmoid function \cite{cadene2019rubi}. Since the bias has been captured by the branch module, the vanilla method has no incentive to learn those superficial correlations. \emph{In the test stage}, the vanilla method is used alone to provide unbiased predictions. 

\fr{Specifically, RUBi \cite{cadene2019rubi}, LMH \cite{clark2019don}, and AdvReg \cite{ramakrishnan2018} leverage a question-only branch with a specific mechanism, such as a sigmoid function, a learned mixin \cite{hinton2002training} strategy, and an adversarial regularization, respectively, to prevent the vanilla method from looking into only one modality. It is noteworthy that this branch is usually implemented by a Recurrent Neural Network (RNN) followed by a multi-layer perceptron. GGE \cite{han2021greedy} decomposes the language bias into two categories, namely distribution bias and shortcut bias. To mitigate them, it employs a greedy gradient ensemble training strategy. The architecture of GGE is also a combination of the question-only branch and the vanilla VQA model $f$.} Inspired by the causal effect, CF \cite{niu2021counterfactual} revisits ensemble-based debiasing methods from a causal inference perspective, formulating the language bias as the direct causal effect of questions on answers, \emph{i.e.}, the pure language effect. It mitigates the bias learning by subtracting the pure language effect from the total causal effect. 

However, the above methods still suffer from the issue of achieving high OOD performance at the expense of ID performance. To address this problem, IntroD \cite{niu2021introspective} introduces a training paradigm that first applies CF as a causal teacher to capture the bias in ID and OOD scenarios, then blends the inductive bias of both worlds fairly, and finally performs distillation for a robust student model. \fr{ Inspired by the above works, CIKD \cite{pan2021distilling} first infers the causal target by exploring counterfactual causality, then employs knowledge distillation \cite{9340578,9737403} to transfer the knowledge from the target, and finally leverages ensemble learning to mitigate the bias. Curriculum learning \cite{bengio2009curriculum} enables models to be trained initially on the easier examples and gradually shift towards the harder ones. Motivated by this, LBCL \cite{lao2021superficial} leverages a visually sensitive coefficient metric to measure the difficulty of each sample and applies an easy-to-hard training strategy to solve the bias learning. D-VQA \cite{wenXTWW21} divides the bias into positive and negative. The former corresponds to inherent rules in the real world, \emph{i.e.}, commonsense knowledge, while the latter is the harmful statistical regularity, \emph{e.g.}, ``tennis" is the answer to most of the ``what sport" questions. From the feature-level view, D-VQA leverages a question-only and image-only branch to capture the uni-modal bias respectively. From the example-level view, it builds negative examples to assist the model training.
}

\subsection{Data Augmentation}
\label{sec:da}
Data augmentation-based methods typically generate additional augmented question-answer pairs $(v_i^{'}, q_i^{'}, a_i^{'})$ for each sample in the original dataset $\mathcal{D}$ to balance the distribution of training data or mitigate the data bias. In particular, the question $q_i^{'}$ is generated using word masking or replacement \cite{9525040,niu2021counterfactual}, the image $v_i^{'}$ is produced by object swapping and mixing \cite{gao2021information}, color conversion \cite{chen2020simple}, and image flipping and resizing \cite{park2023rgb}, and the answer $a_i^{'}$ is obtained depending on the situation. For positive examples, the answer is the same as the ground-truth answer. For counterfactual examples, it is obtained by excluding the top $K$ answers of the corresponding positive example from the answer set. The answer prediction $\hat{a}_i$ process is the same as the paradigm described in Equation (\ref{eq:gp}):
\begin{equation}
    \begin{split}
        & \hat{a}_i = E_{\mathrm{c}} ( E_{\mathrm{m}} ( E_{\mathrm{v}}(v_i), E_{\mathrm{q}}(q_i) ) ) , \\
        & (v_i, q_i, a_i) \in \mathcal{D} \cup \{(v_i^{'}, q_i^{'}, a_i^{'}) | i \in [1,n]\}. 
    \end{split} \label{eq:data}
\end{equation}

\begin{figure}[t]
	\centering  
	\includegraphics[width=\columnwidth]{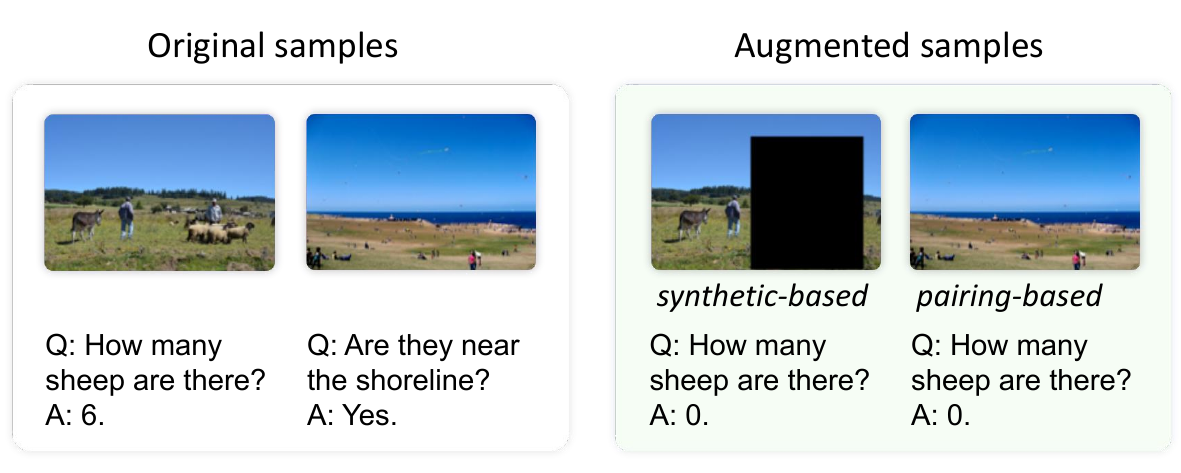}
	\caption{Comparison between original samples and augmented samples. The synthetic-based technique generates new training samples by modifying regions or words of the original images or questions, while the pairing-based technique generates new samples by re-matching relevant questions for images.}
	\label{fig:DA}
\end{figure}

The methods usually perform data augmentation from two perspectives to achieve robust performance: 1) \emph{synthetic-based}: generate new training samples by modifying regions or words of the original images or questions; 2) \emph{pairing-based}: generate new samples by re-matching relevant questions for images. The examples generated by the mentioned techniques are shown in Fig. \ref{fig:DA}. The left is an original sample, while the right is the augmented sample based on the left sample. The synthetic method masks the ``sheep" region in the image, resulting in the original question having a new answer of ``0". Compared with this, the pairing method does not change the original image but rather retrieves questions related to the image.
\begin{figure*}
	\centering  
	\includegraphics[width=\textwidth]{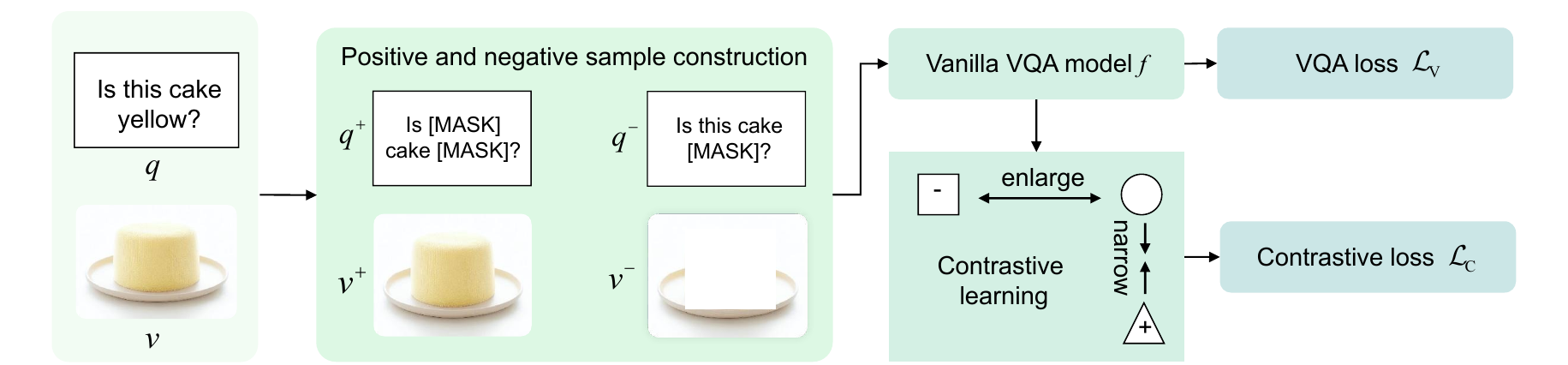}
	\caption{Illustration of self-supervised contrastive learning-based debiasing methods. The contrastive learning loss $\mathcal{L}_{\mathrm{C}}$ is calculated by narrowing the distance between similar samples and enlarging the distance between dissimilar samples in the multi-modal joint embedding space. The model is trained by jointly optimizing VQA and contrastive losses.}
	\label{fig:CL}
\end{figure*}

\noindent\textbf{Synthetic-based.} Kafle \textit{et al.} \cite{kafle2017data} used existing semantic annotations in the dataset and RNN-based methods to produce new questions respectively.
Agarwal \textit{et al.} \cite{agarwal2020towards} leveraged a GAN-based model \cite{shetty2018adversarial} to remove objects and then mitigated the bias by adversarial training. \fr{An ideal VQA model should possess two key characteristics  \cite{chen2020counterfactual}: (1) visual-explainable ability: the model should make accurate predictions by leveraging relevant visual regions. (2) question-sensitive ability: the model should be sensitive to different questions, \emph{i.e.}, the model is expected to yield varying responses for different questions.} To address this issue, CSS \cite{chen2020counterfactual} and ECD \cite{kolling2022efficient} first synthesize counterfactual image-question pairs by masking critical objects in the original image and critical words in the original question and then assign ground-truth answers to those synthesized samples. This can drive the method to employ informative data to answer questions. To force the method to concentrate on the critical elements of inputs, MUTANT \cite{gokhale2020mutant} performs mutating on the input images and questions to expose the model to perceptually similar yet semantically dissimilar samples.

\noindent\textbf{Pairing-based.} To make the synthesized samples more natural, SimpleAug \cite{kil2021discovering} utilizes rich semantic annotations in the training data to pair images with other relevant questions, instead of generating them from scratch and employs a series of rules to ascertain the existence of ground truth answers. However, the reasonable answers generated by the mentioned rules may limit the generalization. To address this issue, KDDAug \cite{chen2022rethinking} relaxes the requirements for reasonable image-question pairs and generates pseudo-answers for all composed pairs using a knowledge distillation-based answer assignment.

\fr{There also exist other alternatives to achieve the purpose of data augmentation. Specifically, DLR \cite{jing2020overcoming} decomposes the question representation into three distinct phrase representations: type, object, and concept, which are then integrated to predict answers. VGQE \cite{kv2020reducing} learns question representations by leveraging both visual information extracted from the image and linguistic information derived from the question before multimodal fusion. CVL \cite{abbasnejad2020counterfactual} uses a causal model with an additional variable to generate counterfactual samples, which compels the VQA model to utilize both input modalities instead of depending on statistical patterns that are specific to either one. The aforementioned methods \cite{agarwal2020towards,chen2020counterfactual,kolling2022efficient,chen2022rethinking} augment data based on the internal (original) data. Compared with this, we can also expand data from external sources. For instance, inspired by meta-learning \cite{finn2017model,huang2018natural,teney2018visual}, ASL \cite{teney2019actively} retrieves the relevant samples with image-question pairs from an external source, which is leveraged to learn adapting parameters for the VQA baseline, such as UpDn \cite{anderson2018bottom}, thus acquiring better generalization ability.}

\subsection{Self-Supervised Contrastive Learning}
Self-supervised contrastive learning \cite{9873966,chuang2020debiased,khosla2020supervised} aims at learning an embedding space where similar sample pairs are positioned closely together while dissimilar ones are widely separated. Its use in robust VQA is still at the starting stage. The paradigm of contrastive learning-based debiasing methods is to first generate positive and negative samples that differ from the original samples using data-augmentation techniques introduced in the above subsection, then perform question answering by the vanilla VQA method $f$ described in Equation (\ref{sec:backg}), and finally be optimized jointly by the contrastive learning loss $\mathcal{L}_{\mathrm{C}}$ of multi-modal representations and the VQA loss $\mathcal{L}_{\mathrm{V}}$, as shown in Fig. \ref{fig:CL}. The joint loss $\mathcal{L} $ can be formulated as follows:
\begin{equation}
    \begin{split}
        & \mathcal{L} = \lambda_{\mathrm{C}} \mathcal{L}_{\mathrm{C}} + \lambda_{\mathrm{V}} \mathcal{L}_{\mathrm{V}}, \\
        & \mathcal{L}_{\mathrm{C}} = \mathop{\mathbb{E}}\limits_{o,p,n \in \mathcal{D}^{*}} \left[-\log\left(\frac{e^{s(o,p)}}{e^{s(o,p)}+e^{s(o,n)}}\right)\right], \\
        & \mathcal{L}_{\mathrm{V}} = - \frac{1}{|\mathcal{D}^{*}|} \sum_{i=1}^{|\mathcal{D}^{*}|} [a_i] \log \hat{a}_i, \\
    \end{split}
\end{equation}
where $\lambda_{\mathrm{C}}$ and $\lambda_\mathrm{V}$ are used to balance contrastive learning and VQA, $s(o,p)$ is the scoring function between the anchor $o$ and the positive sample $p$, $s(o,n)$ is the scoring function between the anchor and the negative sample $n$, $|\mathcal{D}^{*}|$ denotes the number of samples in the augmented dataset such as the dataset described in Equation (\ref{eq:data}), and $[a_i]$ is the index of the answer $a_i$.

Specifically, CSS+CL \cite{liang2020learning} is the first to introduce self-supervised contrastive learning for VQA counterfactual samples, where CSS \cite{chen2020counterfactual} is leveraged to generate factual and counterfactual samples or positive and negative samples. Nevertheless, it has been found that CSS+CL results in a decline in ID performance while only slightly improving OOD performance. To this end, MMBS \cite{si2022towards} attributes the key point of solving language bias to the positive-sample design for excluding spurious correlations, which can boost the OOD performance significantly while retaining the ID performance. It exploits unbiased information through a positive sample construction strategy and employs an algorithm to discriminate between biased and unbiased samples so that they can be handled differently. These methods mitigate training bias from the forward chaining perspective which is similar to the paradigm described in Equation (\ref{eq:gp}), but they rarely explore it from the backward chaining perspective. \fr{Motivated by this, Lao \textit{et al.} \cite{lao2022vqa} introduced a bidirectional chaining framework. In this framework, the forward chaining process resembles the procedure described in Equation \eqref{eq:gp}. On the other hand, the backward chaining process aims to generate crucial visual features by leveraging the annotated answer as a guiding mechanism. Nonetheless, the negative sample generation techniques employed by the aforementioned methods may inadvertently introduce visual shortcut bias \cite{dancette2021beyond}. To tackle this problem, LSP \cite{liu2023flexible} introduces selective sampling rates and question-type-guided sampling, effectively eliminating the reliance of VQA models on visual shortcut bias. Furthermore, drawing inspiration from prompt learning \cite{liu2023pre,zhou2022conditional,xu2023multimodal}, LSP introduces a question-type-guided prompt within the language context, thereby enhancing the significance of questions within the VQA model.}

\subsection{Answer Re-Ranking}
\begin{figure}[t]
	\centering  
	\includegraphics[width=\columnwidth]{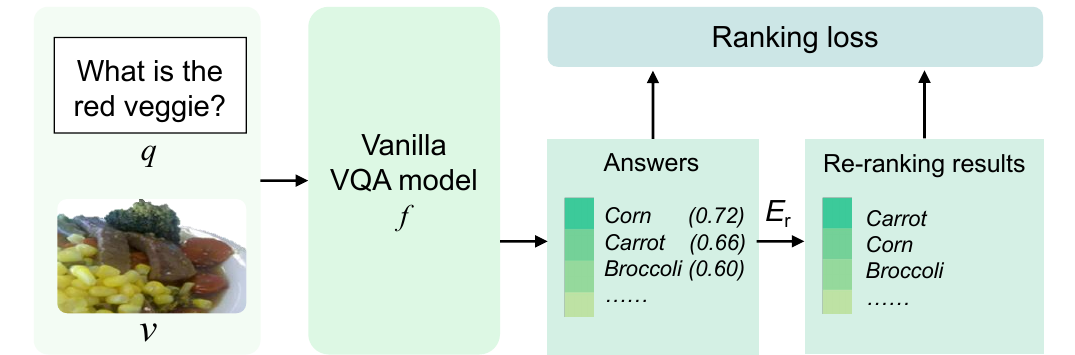}
	\caption{Overview of answer re-ranking-based debiasing methods. The re-ranking module re-ranks the answers predicted by the vanilla VQA model.}
	\label{fig:AR}
\end{figure}

The answer re-ranking-based methods employ the re-ranking mechanism to re-sort the candidate answers provided by vanilla VQA baselines, which can guide the baseline to make better use of visual information. Specifically, their paradigm is to first predict answers using the vanilla method $f$ described in Equation (\ref{sec:backg}), and then re-rank answers leveraging the re-ranking module $E_{\mathrm{r}}$, and finally guide $f$ to provide accurate answers $\hat{a}_i$ by back-propagating the gradient of re-ranking losses, as shown in Fig. \ref{fig:AR}. This process is formulated as follows:
\begin{equation}
    \hat{a}_i  = E_{\mathrm{r}} ( E_{\mathrm{c}} ( E_{\mathrm{m}} ( E_{\mathrm{v}}(v_i), E_{\mathrm{q}}(q_i) ) ) ) . \label{eq:ar}
\end{equation}

Specifically, \fr{GVQA \cite{agrawal2018don} decouples the recognition of visual objects in an image from the identification of plausible answer space for a given question. This is accomplished through the use of a question classifier, which determines the type of questions to reduce the size of answer space, as well as an answer cluster predictor that identifies the expected types of answers, such as object names, colors, and numbers.} SCR \cite{wu2019self} and HINT \cite{selvaraju2019taking} employ a human attention-based penalty mechanism to guide the answer ranking. For each question-answer pair, SCR first determines the region of an image that has the greatest influence on the network prediction of the correct answer. Then, it penalizes the network for concentrating on the region once the prediction is wrong. Similarly, HINT penalizes the model if the pair-wise ranking of visual region sensitivities to ground-truth answers does not match the rankings calculated from human-based attention maps. These two methods made significant progress using the human attention-based penalty mechanism, but Shrestha \textit{et al.} \cite{shrestha2020negative} demonstrated that the performance improvements seen in these methods are due to a regularization effect that prevents overfitting to linguistic priors rather than enhanced visual grounding. Inspired by this, SimpleReg \cite{shrestha2020negative} employs a simple regularization scheme that consistently penalizes the model regardless of whether its predictions are accurate or not.
\begin{figure*}
	\centering  
    \includegraphics[width=\textwidth]{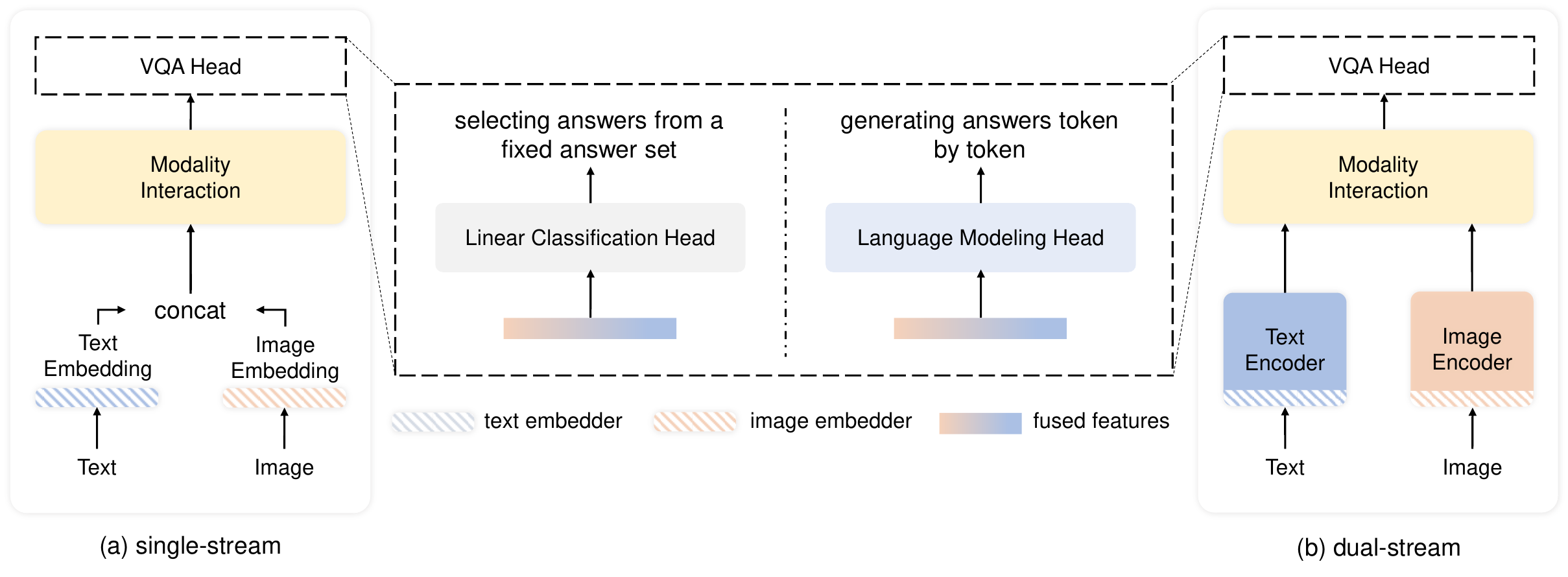}
    \caption{Illustration of single- and dual-stream VLM architectures. The former concatenates the embeddings of texts and images obtained from the embedding layer. The concatenated features are treated as a single embedding sequence and fed into a Transformer block for fusion. On the other hand, dual-stream methods extract features from images and text separately using different encoders. Then, the respective features are treated as different embedding sequences for modality interaction. Both single-stream and dual-stream VLMs can incorporate a linear classification head and a language modeling head to handle discriminative VQA and generative VQA, respectively.}
	\label{fig:VLM}
\end{figure*}

Different from the mentioned methods, RankVQA \cite{qiao2020rankvqa} and SAR \cite{si2021check} explore the combination of answer re-ranking and multimodality tasks to reduce bias learning. They first select candidate answers relevant to the question or the image, and then re-rank the answers by an image caption task \cite{yu2018topic,ye2018attentive,yang2020ensemble} or a visual entailment task \cite{song2022clip,cao2022alignve}, motivated by the idea that the correct answer must be more pertinent to the context of the image than the incorrect answer. These tasks play an important role in verifying whether the image semantically entails or matches the synthetic statement of questions and candidate answers. 

The aforementioned methods mainly focus on strengthening the visual feature learning capability but ignore analyzing its inherent cause and providing an explicit interpretation. Therefore, some work \cite{guo2022loss, lao2021language} suggested taking a look at the robust VQA problem from a class-imbalance perspective. They further demonstrated the effectiveness of the loss re-scaling strategy, which assigns various weights to each answer based on the statistics of the training data to estimate the final loss. \fr{For example, LPF \cite{liang0Z21} leverages a dynamic weighting scheme to each training example and adjusts the VQA loss according to the output distribution of a question-only branch. In addition, there are several works addressing bias learning from the perspective of decision margins. AdVQA \cite{GuoNCJZB21}, for example, employs an adapted margin loss function to distinguish between frequent and sparse answer spaces for each question type. RMLVQA \cite{basu2023rmlvqa} employs an instance-specific adaptive margin loss function to distinguish between hard and easy examples, as well as frequent and rare ones. MFE \cite{gat2020removing} employs a regularization term that relies on a functional entropy, to ensure a balanced contribution of each modality towards classification.}

\section{Exploring the Robustness of Vision-Language Models for VQA}
\label{sec:VLM}
The majority of approaches outlined above are task-specific models that are trained on a restricted amount of data. In contrast, in recent years, Vision-Language Models (VLM) \cite{lu2019vilbert, chen2020uniter, su2019vl, li2019visualbert, tan2019lxmert, kim2021vilt, li2020oscar, radford2021learning, li2021align, bao2022vlmo, wangsimvlm}, which are first pre-trained on large-scale image-text pairs and then fine-tuned on downstream tasks, have become prevalent due to their outstanding performance across a range of multimodal tasks. Recent studies \cite{yu2022coca, li2022blip, wang2023image, chen2022pali, li2023blip} have demonstrated that VLMs are emerging as the mainstream choice for handling VQA. Therefore, it is essential to discuss the robustness of VLMs in this task.

VLMs are often categorized into two classes: single-stream \cite{chen2020uniter, li2020oscar, kim2021vilt} and dual-stream \cite{tan2019lxmert, li2021align, radford2021learning, li2022blip, yu2022coca, wang2023image}. Fig. \ref{fig:VLM} illustrates their architectures adapted to VQA. It can be seen that single-stream methods first concatenate the embeddings of texts and images and then feed the combined features as a single embedding sequence to a Transformer block for fusion. In contrast, dual-stream methods first treat the embeddings of texts and images as distinct sequences and then input them into a Transformer block simultaneously for modality interaction. For the VQA task, single- and dual-stream methods integrate either a linear classification head or a language modeling head after multimodal interaction, depending on the perspective of treating VQA as either a discriminative or generative task.


To provide a deep insight into the mentioned VLM categorization, we choose ViLT and BLIP as the representative works for the above two classes. ViLT \cite{kim2021vilt} is a single-stream VLM that abandons the use of object detectors such as Faster R-CNN, which were previously widely employed by single-stream (UNITER \cite{chen2020uniter}, OSCAR \cite{li2020oscar}) or dual-stream (LXMERT \cite{tan2019lxmert}) VLMs to extract image features on the visual end. Instead, ViLT first splits images into patches and then embeds them simply using a linear projection layer. These image features are then concatenated with word embeddings and fed into a deep Transformer for modality interaction. When dealing with the VQA task, ViLT feeds fused features into a linear classification head for discriminative answer predictions. ViLT's convolution-free architecture significantly reduces model size and inference time, making it tens of times faster. However, the lightweight visual encoding of ViLT leads to a performance decrease on the downstream task that requires salient visual features. BLIP \cite{li2022blip} is a dual-stream VLM that utilizes a multimodal mixture of encoder-decoder architecture, encompassing an image encoder, a text encoder, an image-grounded text encoder, and an image-grounded text decoder. This comprehensive framework enables the integration of vision-language understanding and generation. The method undergoes joint pre-training with image-text contrasting, image-text matching, and language modeling losses, resulting in robust zero-shot generalization capabilities. To handle the VQA task, BLIP inputs image features into the image-grounded text encoder to perform fusion, and the fused features are then fed into the image-grounded text decoder for generative predictions.

\begin{table}[tb]
\centering
\caption{VQA results of VLMs in ID and OOD situations. The symbol PT indicates whether a method has been pre-trained on large-scale datasets. Results marked with $\dagger$, $\ddagger$, $\Diamond$ are reported in \cite{agrawal2022rethinking}, \cite{guo2022loss}, \cite{li2020closer}, respectively. ``DISC" denotes that methods regard VQA as a discriminative task, while ``GEN" represents that methods consider VQA as a generative task. ``Type" denotes the VLM class shown in Fig. \ref{fig:VLM}.} \label{tab:VLM}
\resizebox{\columnwidth}{!}{
\begin{tabular}{rccccc}
\toprule
\textbf{Method}  & \textbf{Type} & \textbf{PT} & \makecell[c]{\textbf{VQA-CP v2} \\ \textbf{test}}  & \makecell[c]{\textbf{VQA v2} \\ \textbf{validation}}  & \textbf{HM} \\     \midrule
UpDn \cite{anderson2018bottom}                                                  & -     & No           & 37.81                   & 65.99                      & 48.07       \\
VILBERT\textsubscript{DISC}\textsuperscript{$\dagger$} \cite{lu2019vilbert}     & (b)   & No           & 42.50                   & 66.70                      & 51.92       \\   
VILBERT\textsubscript{DISC}\textsuperscript{$\dagger$} \cite{lu2019vilbert}     & (b)   & Yes          & 42.90                   & 67.00                      & 52.31       \\   \midrule[0.2pt]
ALBEF\textsubscript{DISC}\textsuperscript{$\dagger$}  \cite{li2021align}        & (b)   & No           & 40.10                   & 64.00                      & 49.31       \\
ALBEF\textsubscript{DISC}\textsuperscript{$\dagger$} (4M) \cite{li2021align}    & (b)   & Yes          & 44.40                   & 70.00                      & 54.34       \\
ALBEF\textsubscript{DISC}\textsuperscript{$\dagger$} (14M)  \cite{li2021align}  & (b)   & Yes          & 45.20                   & 70.30                      & 55.02       \\    \midrule[0.2pt]
ALBEF\textsubscript{GEN}\textsuperscript{$\dagger$}   \cite{li2021align}        & (b)   & No           & 36.60                   & 61.40                      & 45.86       \\
ALBEF\textsubscript{GEN}\textsuperscript{$\dagger$}  (4M) \cite{li2021align}    & (b)   & Yes          & 49.20                   & 71.00                      & 58.12       \\
ALBEF\textsubscript{GEN}\textsuperscript{$\dagger$} (14M) \cite{li2021align}    & (b)   & Yes          & 49.60                   & 72.10                      & 58.77       \\    \midrule[0.2pt]
UNITER\textsubscript{Base}\textsuperscript{$\Diamond$} \cite{chen2020uniter}    & (a)   & Yes          & 46.93                   & 72.70                      & 57.04       \\
UNITER\textsubscript{Large}\textsuperscript{$\Diamond$} \cite{chen2020uniter}   & (a)   & Yes          & 50.98                   & 73.82                      & 60.31       \\     \midrule[0.2pt]
LXMERT\textsuperscript{$\ddagger$} \cite{tan2019lxmert}                         & (b)   & Yes          & 51.78                   & 73.06                      & 60.61       \\  
ViLT \cite{kim2021vilt}                                                         & (a)   & Yes          & -                       & 71.26                      & -           \\
OSCAR \cite{li2020oscar}                                                        & (a)   & Yes          & -                       & 73.61                      & -            \\
CLIP-ViL\cite{shen2022how}                                                      & (b)   & Yes          & -                       & 76.48                      & -             \\
BLIP \cite{li2022blip}                                                          & (b)   & Yes          & 50.71                   & 78.25                      & 61.54          \\
BLIP-2 \cite{li2023blip}                                                        & (b)   & Yes          & -                       & 82.19                      & -         \\
CoCa \cite{yu2022coca}                                                          & (b)   & Yes          & -                       & 82.30                      & -           \\
BEiT-3\textsubscript{Base} \cite{wang2023image}                                                     & (b)   & Yes          & 49.63                   & 77.65                      & 60.56             \\
PaLI \cite{chen2022pali}                                                        & (b)   & Yes          & -                       & 84.30                      & -             \\
PaLI-X \cite{chen2023pali}                                                      & (b)   & Yes          & -                       & 86.00                      & -             \\
\bottomrule
\end{tabular}
}
\end{table}

In recent years, there have been a few works \cite{li2020closer,fang2022data,zhou2022unsupervised, agrawal2022rethinking} to explore the robustness of VLMs dealing with the VQA task. However, overall, research in this area is limited. With the ongoing development of VLMs, further investigation is required to study their robustness facing VQA. We first gather the available results of VLMs on the VQA v2 validation (ID) and the VQA-CP v2 test (OOD) split. Then, we conduct experiments to obtain the performance of BLIP and BEiT-3 on the VQA-CP test split. The results are shown in Table \ref{tab:VLM}. From this table, we can observe the following phenomena. Firstly, although VLMs exhibit excellent performance in the ID scenario, their accuracy significantly drops in the OOD scenario. For example, the performance of UNITER\textsubscript{Base} drops from 72.20\% to 46.93\%. Secondly, from the perspective of model paradigms (discriminative or generative), generative models seem to be more robust than discriminative models. For instance, ALBEF\textsubscript{GEN} (4M) outperforms ALBEF\textsubscript{DISC} (4M) by 4.8\% and 1.0\% on the mentioned split, respectively. Thirdly, considering the model's parameter size, larger models appear to achieve better results in both ID and OOD scenarios. Specifically, it can be seen that ALBEF\textsubscript{DISC} (14M) is superior to ALBEF\textsubscript{DISC} (4M) by 0.8\% and 0.3\% on ID and OOD situations, respectively. Finally, models with pre-training on large-scale datasets are more robust than those without pre-training. For example, the HM accuracy of ALBEF\textsubscript{DISC} with pre-training is superior to that without pre-training by 12.26\%. \cite{jolly2020can} also points out that VQA models with pre-trained text encoders are more robust to lexical variations of input questions. Additionally, the inherent flaws in both the VQA datasets and the evaluation metric significantly influence the performance of VLMs, which we will discuss further in Section \ref{sec:dis}.


\section{Discussions and Future Directions}
\label{sec:dis}
Based on the comprehensive analysis of existing datasets, evaluations, and methods outlined above, it becomes apparent that there is potential for improvement in robust VQA. Consequently, our focus will now shift toward discussing the strategies and areas where future advancements can be made.
\begin{table*}[t]
\centering
\caption{The accuracy (\%) comparison of non-debiasing (ND) and debiasing (DE) methods on various datasets. All the results are on the test split except for the result of the VQA v2.0 testdev split.}
\label{tab:methods-comp}
\resizebox{\textwidth}{!}{
\begin{tabular}{cccccccccccccccccc}
\toprule
\multirow{2}{*}{\textbf{Type}} & \multirow{2}{*}{\textbf{Backbone}} & \multirow{2}{*}{\textbf{Methods}} & \multicolumn{4}{c}{\textbf{VQA v2.0 test-dev}}               & \multicolumn{4}{c}{\textbf{VQA-CP v1 test}}                  & \multicolumn{4}{c}{\textbf{VQA-CP v2 test}}                  & \multicolumn{3}{c}{\textbf{GQA-OOD val}}     \\ \cmidrule(l){4-7} \cmidrule(l){8-11} \cmidrule(l){12-15} \cmidrule(l){16-18}
                               &                                    &                                   & \textbf{All} & \textbf{Y/N} & \textbf{Num.} & \textbf{Other} & \textbf{All} & \textbf{Y/N} & \textbf{Num.} & \textbf{Other} & \textbf{All} & \textbf{Y/N} & \textbf{Num.} & \textbf{Other} & \textbf{All} & \textbf{Tail} & \textbf{Head} \\ \midrule
\multirow{2}{*}{NDE}           & \multirow{2}{*}{NA}                & SMRL                              & 64.76        & 82.20        & 46.44         & 54.01          & 36.86        & 43.39        & 12.88         & 40.22          & 37.09        & 41.85        & 12.76         & 41.28          & 46.32        & 41.67         & 49.16         \\
                               &                                    & UpDn                              & 65.78        & 83.07        & 45.88         & 55.54          & 37.40        & 43.27        & 12.89         & 41.57          & 38.04        & 43.41        & 12.92         & 42.26          & 47.75        & 42.62         & 50.89         \\ \midrule
\multirow{6}{*}{DE}            & \multirow{3}{*}{SMRL}              & CF Variant                        & 62.63        & 82.14        & 44.02         & 50.14          & 43.76        & 60.83        & 13.92         & 38.92          & 54.04        & 88.23        & 30.86         & 42.71          & 39.34        & 35.09         & 41.95         \\
                               &                                    & RUBi                              & 63.28        & 82.28        & 45.46         & 51.05          & 50.83        & 80.18        & 16.52         & 39.43          & 47.61        & 74.68        & 20.31         & 43.23          & 46.78        & 42.52         & 49.39         \\
                               &                                    & CF                                & 63.01        & 81.96        & 45.48         & 50.98          & 56.88        & 89.75        & 17.56         & 40.21          & 55.42        & 90.56        & 26.61         & 45.65          & 44.28        & 41.20         & 44.28         \\ \cmidrule(l){2-18} 
                               & \multirow{3}{*}{UpDn}              & CF Variant                        & 65.19        & 82.98        & 44.93         & 54.58          & 37.26        & 44.99        & 13.08         & 41.68          & 37.59        & 44.04        & 13.03         & 41.97          & 48.03        & 44.21         & 50.38         \\
                               &                                    & RUBi                              & 64.94        & 83.22        & 45.51         & 53.71          & 50.45        & 80.25        & 14.76         & 41.01          & 39.57        & 49.74        & 19.17         & 42.38          & 48.03        & 42.24         & 51.59         \\
                               &                                    & CF                                & 65.47        & 83.16        & 44.72         & 55.07          & 57.64        & 89.18        & 14.57         & 43.75          & 54.02        & 91.35        & 13.46         & 45.60          & 45.24        & 41.11         & 47.78         \\ \bottomrule
\end{tabular}
}
\end{table*}

\noindent\textbf{Does the dataset annotation exhibit a high level of quality?} Most of the existing datasets were developed based on the VQA v2 dataset \cite{goyal2017making}. However, the answer annotations in the dataset often lack consistent agreement, which can result in inaccurate evaluation outcomes. For instance, as illustrated in Fig. \ref{fig:vqa-v1-eg}, a system that generates a ``Yes" or ``No" answer for the ``Yes/No" question would receive an accuracy score of ``1.00" or ``0.67", respectively. Therefore, current data quality cannot support the accurate performance measurement of VQA models. In the future, we should ensure the quality of data annotations \cite{monarch2021human,jollyetal2021ease}, such as introducing a process where annotations are reviewed by experienced annotators or experts for ambiguous cases.

\noindent\textbf{What datasets should be developed?} Although existing datasets, especially the OOD dataset, enable us to provide insight into the robustness of VQA methods \cite{malinowski2017ask,wang2022revise}, it is essential to note that each dataset has its unique limitations. Taking the most commonly used OOD dataset VQA-CP \cite{agrawal2018don} as an example, it has two shortcomings. First, its distribution between training and test splits is significantly different or even reversed, which may not align with the real-world scenario. Some methods \cite{agrawal2018don,cadene2019rubi,lao2021superficial} may be devised based on this prior, which may not reflect the robustness of these methods accurately. Second, VQA-CP lacks the validation split, which results in methods being tuned on the test split. Although GQA-OOD \cite{kervadec2021roses} alleviates the above issues, its test split is too small, only containing 12,578 questions. \emph{Therefore, existing datasets may not be sufficient to evaluate robustness. Furthermore, the dataset does not involve fine-grained bias evaluations such as vision shortcut measurement. To address this issue, we should develop a dataset that satisfies the following properties in the future.} 
\begin{itemize}
    \item The dataset should be sufficiently large and complete, with adequate validation splits for fine-tuning hyper-parameters and large splits for training and testing. 
    \item The dataset should contain ID and OOD test settings simultaneously. In this way, we can conduct a comprehensive and fair evaluation of the robustness of VQA methods.
    \item The distribution between training and test splits should be more natural, rather than artificially setting significantly different or even contradictory data distributions. The artificial distribution prior may be used to improve model performance, while the prior can not be applied to other situations, leading to poor generalization ability.
    \item The OOD test setting should simultaneously include language, vision, and multimodality bias, to have a more refined assessment of robustness.
    \item The question format should be various, particularly in the test split. The question in existing datasets is usually generated by the template. However, the question patterns generated by templates may be not enough, and be learned or memorized easily, resulting in inaccurate comparisons.
\end{itemize}

\noindent\textbf{Are the evaluation metrics effective enough?} Current evaluation protocols assign equal weight to each question. However, some questions, such as the OOD question requiring multi-hop reasoning, should be treated as more important. Therefore, in the future, we should devise an evaluation protocol that can assign different weights to questions according to annotations, such as the distribution they belong to and their difficulty. With the advancements in VLMs, we observe a growing trend in robust VQA studies using decoder-based architectures \cite{wangofa,xumi23} to generate answers. They may generate answers that include the critical information of annotated answers but have a few additional supplementary or different details. For example, given a question ``What is the child doing in this picture?", the model may generate an answer like ``The child is dribbling over", while the annotated answer may be ``playing basketball". In this case, the prediction is regarded as a wrong answer by the standard accuracy. Therefore, a comprehensive evaluation metric that can deal with multiple complex situations needs to be explored and proposed. For example, we may apply composite metrics such as accuracy, similarity, and GPT score \cite{liu2303g} to evaluate VQA methods from various angles.

\noindent\textbf{Are the existing debiasing methods robust enough?} A variety of debiasing methods \cite{liang2020learning,li2021adversarial,gupta2022swapmix} for VQA have been proposed and have achieved significant success on VQA v2 and VQA-CP which are the most commonly used datasets to evaluate robustness. This inspires us to consider whether they are robust to other datasets, such as GQA-OOD \cite{shrestha2022investigation}. To address our curiosity, we select several non-debiasing methods, including SMRL \cite{cadene2019murel} and UpDn \cite{anderson2018bottom}, and debiasing methods, including RUBi \cite{cadene2019rubi} and CF \cite{niu2021counterfactual}, and then conduct experiments on these three datasets simultaneously for a fair comparison. The results are shown in Table \ref{tab:methods-comp}. It can be seen that existing debiasing methods are backbone-sensitive and cannot achieve ongoing success on other datasets. \fr{For instance, when using the CF Variant with SMRL, an accuracy of 43.76\% is achieved on the VQA-CP v1 dataset. In contrast, when employing the CF variant with UpDn, the accuracy drops to 37.26\%. Additionally, CF with UpDn attains an accuracy of 45.24\% on the GQA-OOD validation split, which is 2.51\% lower than that of UpDn alone.} We can also note that current debiasing methods still achieve high OOD performance at the expense of ID performance. Specifically, RUBi with SMRL obtains the improvement of 0.46\% on the GQA-OOD validation split, while it drops by 0.48\% on the VQA v2.0 test-dev split. Moreover, existing methods cannot give up answering questions in uncertain situations like humans, just like humans say ``I don't know", but rather predict a wrong answer \cite{whitehead2022reliable}. These show that we should develop true robust or reliable methods in the future to perform well in a variety of ID and OOD performances at the same time.

\noindent\textbf{Are the existing VLM-based VQA methods robust enough?} In recent years, VLMs have gained significant attention and achieved remarkable success across various tasks \cite{mogadala2021trends}, establishing themselves as a prominent research area. Table \ref{tab:VLM} presents the performance of representative VLMs on the VQA-CP v2 test split and the VQA v2 validation split. Notably, early VLMs like UNITER and LXMERT exhibit a certain degree of robustness, but there remains room for improvement, especially in reducing the performance gap between ID and OOD situations. Moreover, there appears to be limited research exploring the robustness of more recent VLMs such as PaLI and PaLI-X on VQA-CP and GQA-OOD. We think that exploring these models in the continual learning setting \cite{nikandrou2022task} could be an intriguing direction for VQA robustness. In this configuration, VLMs can accumulate knowledge and skills through continual learning experiences, thereby facilitating their adaptation to dynamic datasets and evolving scenarios. Furthermore, considering the debiasing techniques, we think data augmentation may be the most perfect integration technology with VLMs due to its model-agnostic characteristics. It is well known that VLMs are usually pre-trained on large amounts of image-text pairs. We think the augmented samples introduced in Section \ref{sec:da} can be added during the pre-training stage. This can enhance the visual grounding capability of VLMs, \emph{i.e.}, robustness.

\noindent\textbf{How should we verify the robustness of the method?} Existing works frequently employ VQA v2 and VQA-CP to assess both debiasing methods and VLMs. These methods, however, do not perform well on the other OOD datasets, as shown in Table \ref{tab:methods-comp}. This demonstrates that it is insufficient to evaluate the robustness only on one OOD dataset. Therefore, considering current dataset conditions, we should leverage multiple ID and OOD datasets to verify the robustness simultaneously, which can reflect the robustness more comprehensively and accurately.

\section{Conclusion}
\label{sec:con}
This paper presents a comprehensive survey that focuses on the domain of robust Visual Question Answering (VQA).  We conduct a systematic review of existing datasets from ID and OOD angles, evaluation metrics from the single and composite views, and methods from the perspective of debiasing techniques. Specifically, we classify the existing debiasing methods into four categories: ensemble learning, data augmentation, self-supervised contrastive learning, and answer re-ranking. We review the robustness of vision-and-language pre-training methods on VQA, classifying them into four categories according to the relative computational size of text encoders, image encoders, and modality interaction modules. We discuss future research directions that should be prioritized for robust VQA.

\ifCLASSOPTIONcompsoc
  \section*{Acknowledgments}
\else
  \section*{Acknowledgment}
\fi

This work was supported by the National Key Research and Development Program of China (2021YFB1715600), the National Natural Science Foundation of China (62137002, U22B2019, 62306229, 62272372, 62293553, 62250066, 62202369), the Natural Science Basic Research Program of Shaanxi (2023-JC-YB-593), the Youth Innovation Team of Shaanxi Universities ``Multi-modal Data Mining and Fusion'', and the Shaanxi Undergraduate and Higher Education Teaching Reform Research Program (23BY195).

\bibliographystyle{IEEEtran}
\bibliography{reference}

\ifCLASSOPTIONcaptionsoff
  \newpage
\fi
\vspace{5pt}

\begin{IEEEbiography}[{\includegraphics[width=1in,height=1.25in,clip,keepaspectratio]{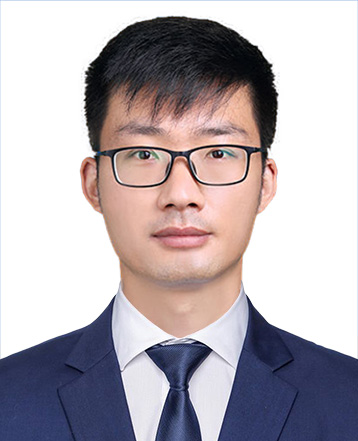}}]{Jie Ma}	
(Member, IEEE) is an Assistant Professor in the School of Cyber Science and Engineering (Faculty of Electronic and Information Engineering) at Xi'an Jiaotong University, Xi'an, Shaanxi 710049, P.R. China. He is also a member of the Ministry of Education's Key Lab for Intelligent Networks and Network Security. His research interests cover natural language processing and trustworthy multimodality learning, focusing particularly on knowledge graph learning, robust visual question answering, and question dialogue. He has contributed to several top journals and conferences, including IJCV, IEEE TIP, TNNLS, IJCAI, and WSDM. Furthermore, he has served as a program committee member for numerous conferences such as ICLR and AAAI, and reviewed manuscripts for multiple journals such as IEEE TIP and TNNLS. For more information, please visit \url{https://dr-majie.github.io/}.
\end{IEEEbiography}
\begin{IEEEbiography}[{\includegraphics[width=1in,height=1.25in,clip,keepaspectratio]{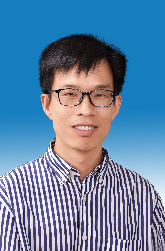}}]{Pinghui Wang}
	(Senior Member, IEEE) is currently a professor with the MOE Key Laboratory for Intelligent Networks and Network Security, Xi’an Jiaotong University, Xi’an, China. His research interests include internet traffic measurement and modeling, traffic classification, abnormal detection, and online social network measurement.
\end{IEEEbiography}

\begin{IEEEbiography}[{\includegraphics[width=1in,height=1.25in,clip,keepaspectratio]{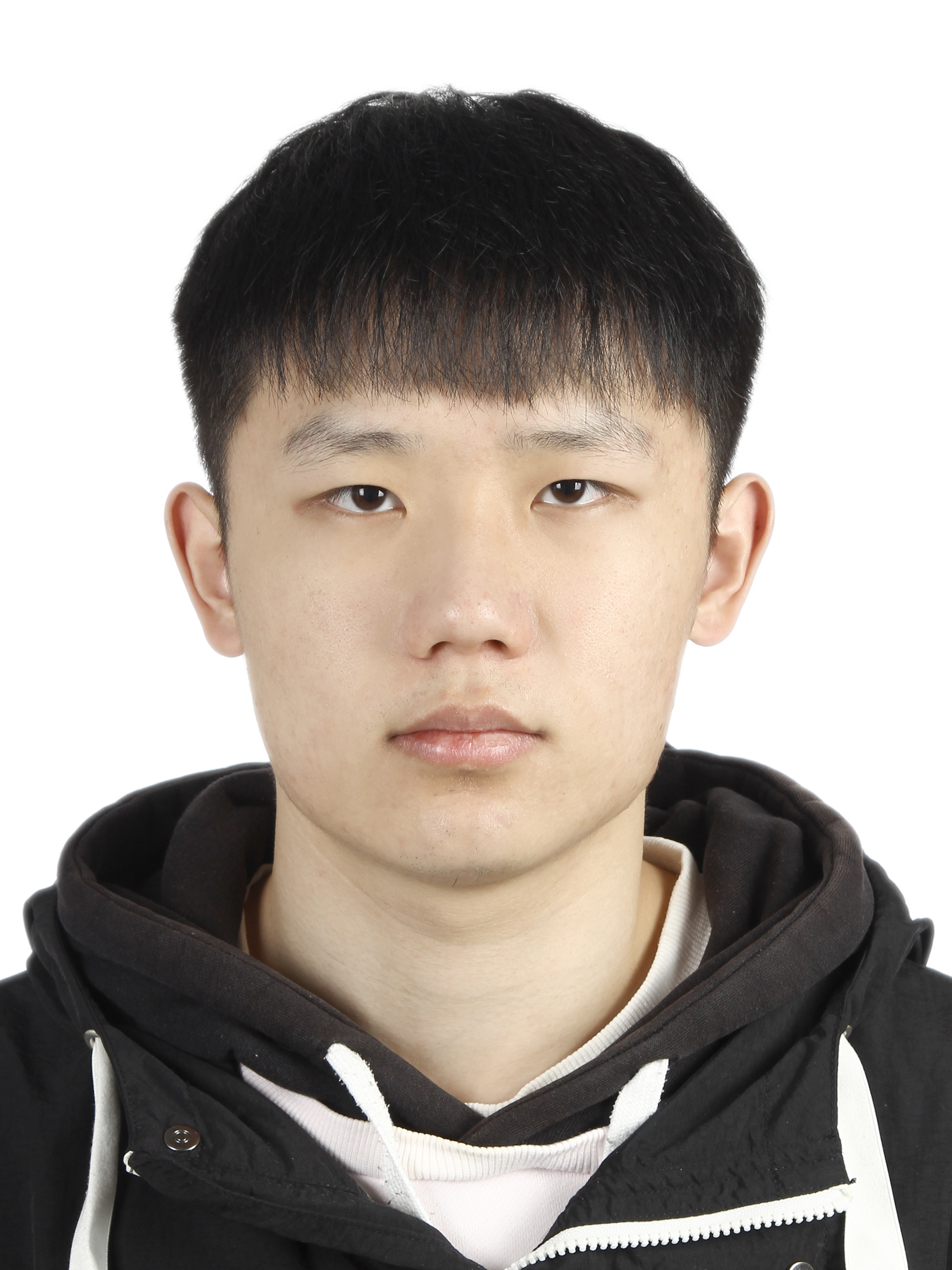}}]{Dechen Kong}
    received the BE degree in Electrical Engineering and Automation from Xi'an Jiaotong University, China, in 2022. He is currently working toward an ME degree in Electronic and Information Engineering at Xi'an Jiaotong University. His research interests include multimodal learning, natural language processing, and computer vision.
\end{IEEEbiography}

\begin{IEEEbiography}[{\includegraphics[width=1in,height=1.25in,clip,keepaspectratio]{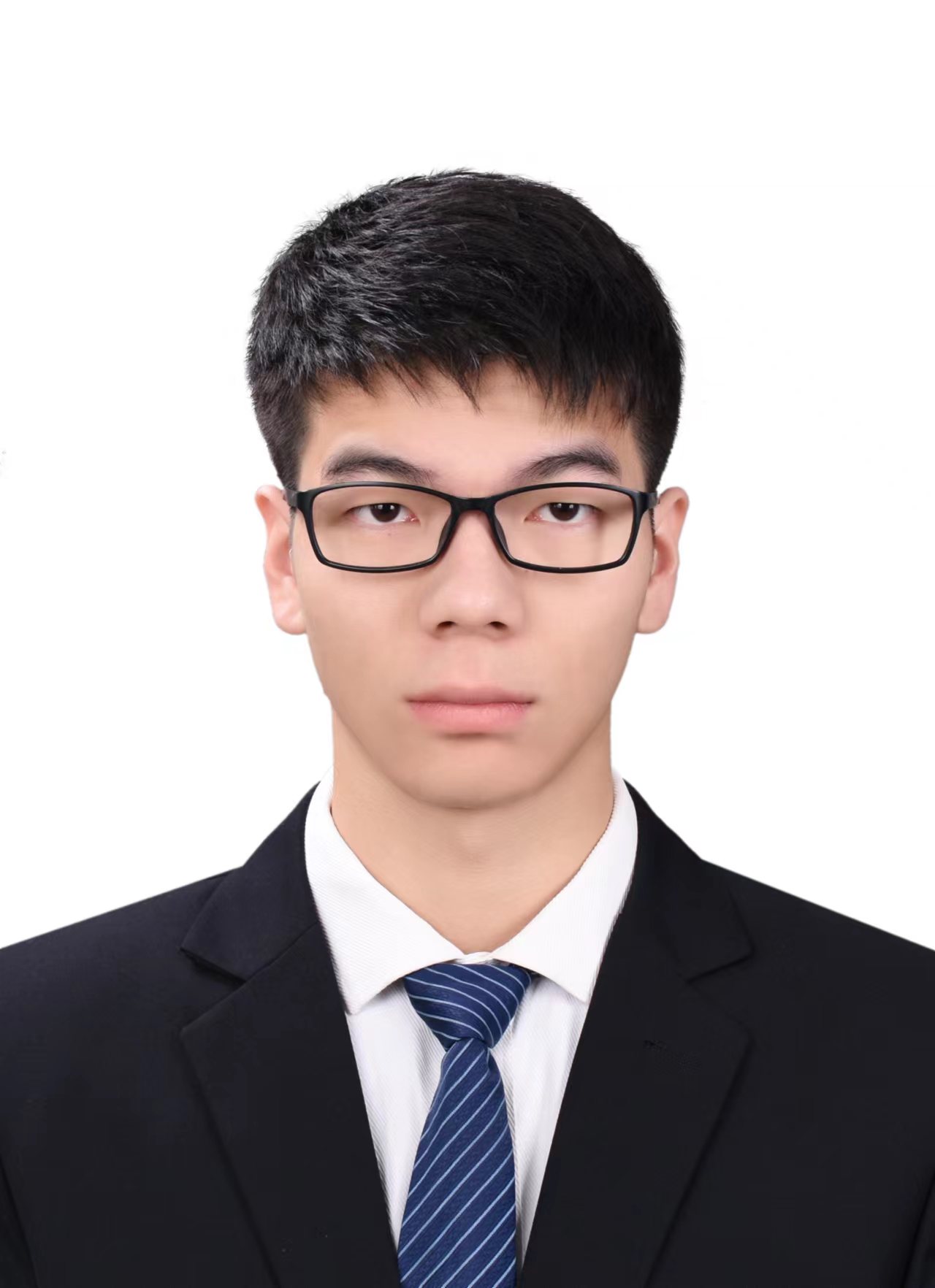}}]{Zewei Wang}
    received the BE degree in automation from Xi'an Jiaotong University, China, in 2022. He is currently working toward the ME degree in control science and engineering with the School of Electronic and Information Engineering, Xi'an Jiaotong University. His research interests include multimodal learning, knowledge graph learning, and visual question answering.
\end{IEEEbiography}

\begin{IEEEbiography}[{\includegraphics[width=1in,height=1.25in,clip,keepaspectratio]{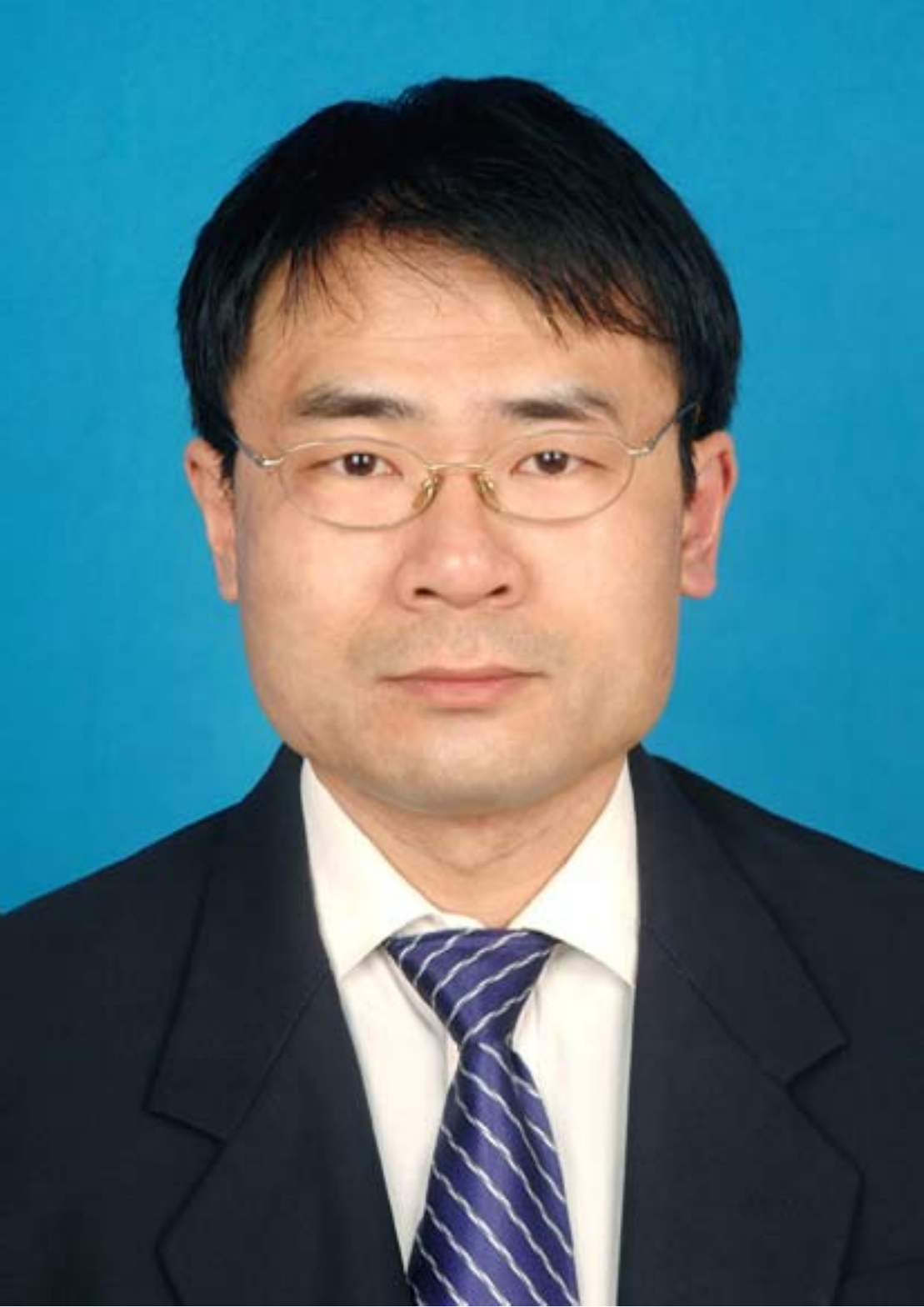}}]{Jun Liu}
	(Senior Member, IEEE) received the B.S. in computer science and technology in 1995 and the Ph.D. degree in systems engineering in 2004, both from Xi’an Jiaotong University, China. He is currently a Professor with the Department of Computer Science, at Xi’an Jiaotong University. He has authored more than ninety research papers in various journals and conference proceedings. He has won the best paper awards in IEEE ISSRE 2016 and IEEE ICBK 2016. His research interests include NLP and e-learning. Dr. Liu currently has served as an associate editor of IEEE TNNLS since 2020 and served as a guest editor for many technical journals, such as Information Fusion and IEEE SYSTEMS JOURNAL. He also acted as a conference/workshop/track chair at numerous conferences.
\end{IEEEbiography}

\begin{IEEEbiography}[{\includegraphics[width=1in,height=1.25in,clip,keepaspectratio]{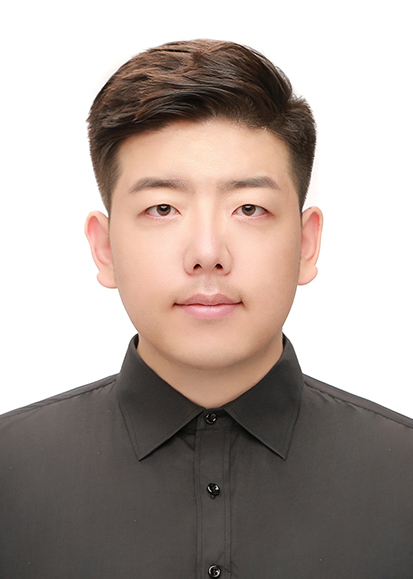}}]{Hongbin Pei}
	is currently an Assistant Professor in the School of Cyber Science and Engineering at Xi’an Jiaotong University, China. He received his B.S., M.S., and Ph.D. degrees from Jilin University, China, in 2012, 2015, and 2021, respectively. His research interests include Deep Learning, Graph Data Mining, and Data-Driven Complex System Modeling, with Applications to Chemistry and Public Health.
\end{IEEEbiography}

\begin{IEEEbiography}[{\includegraphics[width=1in,height=1.25in,clip,keepaspectratio]{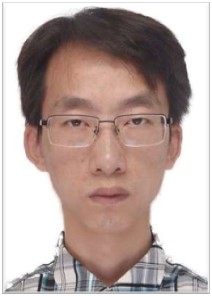}}]{Junzhou Zhao}
    received BS and PhD degrees in control science and engineering from Xi’an Jiaotong University, in 2008 and 2015, respectively. He is currently an associate professor with the School of Cyber Science and Engineering, at Xi’an Jiaotong University. His research interests include graph data mining and streaming data processing.
\end{IEEEbiography}


\vfill

\end{document}